%% file: manu.tex
\documentclass[onecolumn]{IEEEtran}
\IEEEoverridecommandlockouts
\usepackage{cite}
\usepackage{psfig}
\usepackage{amsmath,amssymb,amsfonts,bm}
\usepackage{lscape}
\usepackage{multirow, booktabs}
\usepackage{float}
\usepackage{graphicx}
\usepackage{textcomp}
\usepackage{xcolor}

\usepackage{subfigure}
\usepackage[ruled,vlined]{algorithm2e}
\newcommand{\includeImgPrefix}[2]{\includegraphics[#1]{#2}}
\newcommand{\datasetsubfig}[1]{	\subfigure{\resizebox*{1.7cm}{!}{
	 		\includeImgPrefix{scale=1}{#1}}}}
\newcommand{\expfirstcmp}[1]{	\subfigure{\resizebox*{3.8cm}{!}{
	 		\includeImgPrefix{scale=1}{#1}}}}

\newcommand{\expsecdcmp}[1]{\subfigure{\resizebox*{4.2cm}{!}{
	 		\includeImgPrefix{scale=1}{#1}}}}

\begin{document}
	\input{title/title}
	\input{abstract/abstract}

	\input{introduction/introduction}	
	\input{relatedWorks/relatedworks}

	\input{proposedMethod/proposedMethod}
	\input{experiments/experiments}

	\input{conclusion/conclusion}	
\bibliography{paper}
\end{document}

%% file: title/title.tex

\title{Learning Orientation-Estimation Convolutional Neural Network for Building Detection in Optical Remote Sensing Image}

\author{\IEEEauthorblockN{Yongliang Chen}\\
	
	\IEEEauthorblockA{\textit{Key
			Lab of Optoelectronic Technology and System of Education Ministry} \\
		\textit{Chongqing University}\\
		Chonqing, China}\\
	\IEEEauthorblockA{
		ylchen@cqu.edu.cn}
	}
\maketitle

%% file: abstract/abstract.tex
\begin{abstract}
Benefiting from the great success of deep learning in computer vision, CNN-based object detection 
methods have drawn significant attentions. Various frameworks have been proposed which show awesome and robust
performance for a large range of datasets. 
However, for building detection in remote sensing images, buildings always pose a diversity
of orientations which makes it a challenge for the application of
 off-the-shelf methods to building detection.
In this work, we aim to integrate orientation regression into the popular axis-aligned bounding-box detection method
to tackle this problem.
 To adapt the axis-aligned bounding boxes to arbitrarily orientated ones,
  we also develop an algorithm to estimate the Intersection over Union
(IoU) overlap between any two arbitrarily oriented boxes which is convenient to implement 
in Graphics Processing Unit (GPU) for accelerating computation.
 The
proposed method utilizes CNN for both robust feature extraction and rotated bounding box regression.
 We present our model in an end-to-end fashion making it easy to train. 
The model is formulated and trained to predict orientation, location and extent simultaneously obtaining
tighter bounding box and hence, higher mean average precision (mAP).
 Experiments on remote sensing images of different scales
shows a promising performance over the conventional one.

\end{abstract}
\
\begin{IEEEkeywords}
	building detection, remote sensing image, convolutional neural network, orientation regression, intersection over union estimation
\end{IEEEkeywords}

%% file: introduction/introduction.tex
\section{Introduction}
With building being a key factor in region management and
planning, building detection in very high resolution (VHR)
optical remote sensing images, which has numerous applications
such as damage assessments by comparison of detection results \cite{brunner_earthquake_2010}
and map updating \cite{bonnefon_geographic_2002},
 becomes one of the inevitable challenge
in aerial and satellite image analysis. Extensive methods have
been exploited by researchers in recent years. 
Generally, those methods can be classified into four
categories\cite{cheng_survey_2016}: 
template matching-based methods\cite{stankov_detection_2014},
knowledge-based methods\cite{ahmadi_automatic_2010},
OBIA (object-based image analysis)-based methods\cite{shi_accurate_2015},
 and machine learning-based methods\cite{akcay_automatic_2016}.

 Thanks to the advancing in feature representations and classifier
 design, machine learning-based methods has
 drawn significant attention of researchers, which casts detection
 work into a classification task. In \cite{han_object_2015},
  lower-level features containing spatial and structural information is extracted to
 construct higher features, which is followed by a Bayesian
 framework to learn object detector iteratively. \cite{cheng_object_2013} build HOG (Histogram of Oriented)-based feature pyramids to train a
 support vector machine (SVM). \cite{cohen_rapid_2016} exploit the complex patterns
 of contrast features provided by the training data to model
 buildings, in which detection problem is also reduced to a
 machine learning classification task by introducing candidate
 region search.

 More recently, deep learning \cite{krizhevsky_imagenet_2012}
  has made a significant breakthrough in the field of computer vision. Convolutional
 neural networks (CNNs) with deep structure, which directly
 processes the raw pixels of input image to generate multiple level
 feature representations with semantic abstracting properties, has
 shown an impressively strong power in feature representation and
 obtained a series of success in a wide range of applications\cite{lecun_deep_2015}.
  As deep CNN becomes a powerful tool in feature extraction, more
 sophisticated detection methodologies have been developed
 \cite{girshick_fast_2015,girshick_rich_2014,he_mask_2017,liu_ssd:_2016,redmon_you_2016,ren_faster_2015}.
  In these work, feature maps gathered from convolutional layers
 play a key role. 
 \cite{girshick_rich_2014} use CNNs to extract
 features from each candidate region for the classifier.
 \cite{girshick_fast_2015} does
 classification directly on the feature maps by introducing ROI
 (region of interest) layer, which significant reduces computation
 cost, and applies a regression layer for finer locating objects.
 
 \begin{figure}
 		\centering
 		\subfigure[]{\resizebox*{7cm}{!}{
 		\includeImgPrefix{width=\textwidth}{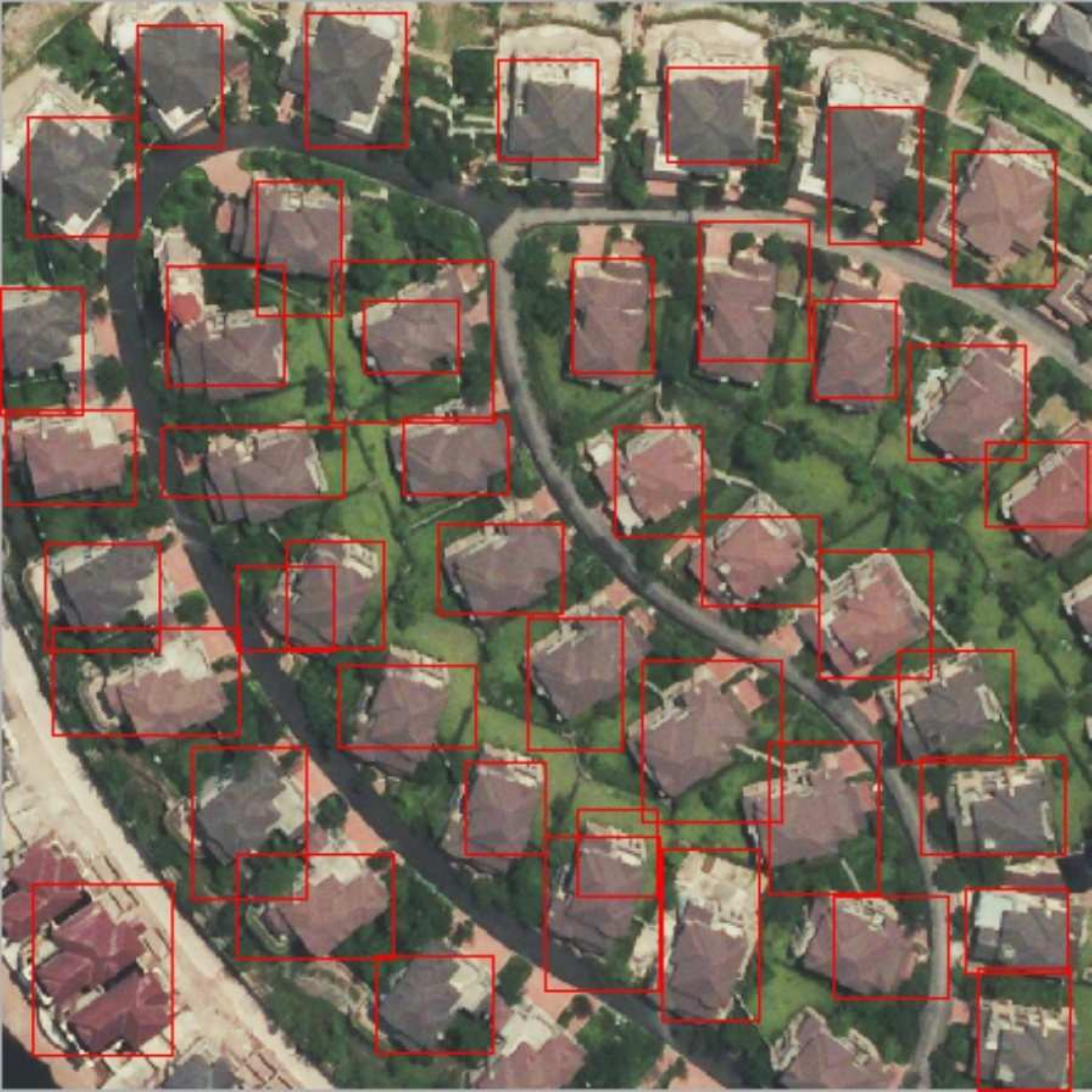}}}
		\subfigure[]{\resizebox*{7cm}{!}{
 		\includeImgPrefix{width=\textwidth}{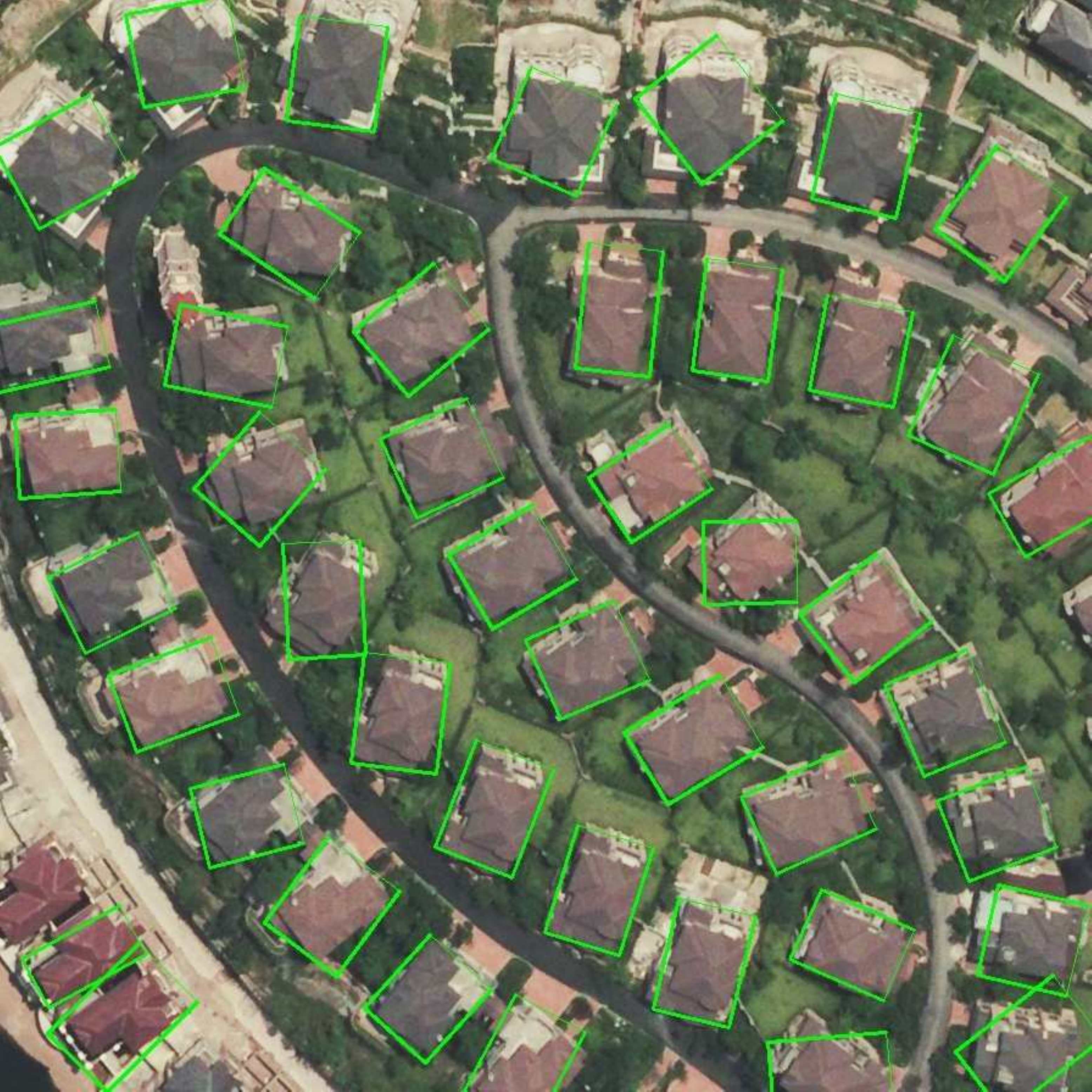}}}
 	\caption{An intuitive comparison between axis-aligned detection
 		(a) and our proposed method (b).
 		 Our method leads to tighter
 		and more precise bounding box for each object, especially when
 		processing image shot with an arbitrary rotation which is
 		commonly observed in optical remote sensing.
 	}\label{fig:intuitive comparison}
 \end{figure}

 To deal with heavily computational budget and fully utilize
 computation capability of graphic processing unit (GPU), 
\cite{ren_faster_2015}
 integrate generating candidate regions into the networks by
 designing proposes region proposal network (RPN) which
 coarsely locates objects via predicting class and offset for each
 anchor box, which is configured by a set of hyperparameters.
 Having achieved attractive performance in both speed and
 accuracy, the idea of detecting on feature map becomes popular. \cite{redmon_you_2016} frame detection as a regression problem and
 simplifies the pipeline by replacing the whole system with a
 single neural network. \cite{liu_ssd:_2016}
  utilize multiple feature layers to
 improve accuracy while keeps time cost low.

CNN is naturally applicable to VHR optical remote sensing
 image, particularly the RGB aerial image. 
 \cite{wang_road_2015} combine both deep
 CNN and finite state machine to extract road networks from VHR
 aerial and satellite imagery. \cite{langkvist_classification_2016}
  use CNN for per-pixel
 classification to improve high-level segmentation. \cite{li_vehicle_2016} work
 towards to a CNN-based method for vehicle detection. Similarly,
\cite{nemoto_building_2017} employ also exploits deep CNNs to effectively extract
 features to improve building detection.
 
 However, different from natural scene images, where objects
 are typically in an upright orientation and thus could be
 reasonably described by axis-aligned bounding boxes, objects in
 optical remote sensing images usually pose a diversity of
 orientations which hampers the use of the off-the-shelf
 methods. To deal with this problem, \cite{cheng_learning_2016}
  propose a rotation-invariant CNN (RICNN) model to advance the performance of
  object detection in these images where objects with rotation
  variation dominate the scope. \cite{chen_learning_2017}
   propose ORCNN (Oriented R-CNN)
    to tackles this problem by applying an additional classifier
  on the detected regions to get orientations from the six
  predefined angle classes.
  \cite{li_rotation-insensitive_2017}
  exploit RPN by adding
  multiangle anchor boxes besides the conventional ones to address
  rotation variations and appearance ambiguity.

  In this work, we present a novel method derived from ORCNN
  and RPN, which predicts building’s orientation in a regression
  fashion. The model detects
  buildings in optical remote sensing images with not only locating their
  pixel coordinates and sizes but also providing their orientation
  information in the format of oriented bounding box as 
  \figurename{\ref{fig:intuitive comparison}} shows.
  
We summarize our contributions as follows:
  \begin{itemize}
  \item We propose a novel method derived from RPN which not
  only locates buildings in optical remote sensing image with
  bounding boxes but also provides orientation information by
  regressing their orientation angles.
  \item A numerical algorithm is also developed for fast
  estimating Intersection over Union (IoU) between two
  arbitrarily given rectangles, which is capable of being
  implemented in GPU allowing accelerated computation.
  \end{itemize}

  In Section \ref{sec:realted_work}, we review some related works and, by comparing
  then with ours we shed light on innovations in this work. Section
  \ref{sec:proposed_method} describes the proposed model in detail. Experiments are
  conducted in Section \ref{sec:experiments},
   followed by conclusions in Section \ref{sec:conclusion}.
   
%
%
   

%% file: relatedWorks/relatedworks.tex
\section{Related Work}\label{sec:realted_work}

There are some methods involving more sophisticated data
to get more accurate estimation which have the ability to
handle more complex building shapes. As \cite{awrangjeb_using_2016}
shows, the author takes point cloud data as the input to extract
precise edges. However, point cloud data is dense and requires
more efforts to obtain which also poses challenges for processing.
In most cases, we use RGB-channel image as input.

In \cite{ren_faster_2015}, 
the system firstly choose a number of candidate
regions from a series of pre-configured bounding boxes, namely
the anchors, by coarsely identifying whether an object is
captured by any anchor. This stage
is performed within RPN. And then, further classification and
regression for each cropped region are simultaneously carried out. In this
stage, object is classified into a certain class and a precise
relative location is also regressed. For  effectively
leveraging the extracted features, all predictions are 
performed on the feature maps.

To estimate the orientation angles, \cite{chen_learning_2017} exploit
the detection result by partitioning the
radius space into 6 bins (i.e., {30, 60, 90, 120, 150 deg}) and
selecting one to describe the orientation for each finally detected region.
In our works, we propose angle anchor box to take orientation
into account and by extending the framework of \cite{ren_faster_2015}
 we cast the
task into a regression problem. 
To achieve this goal, a numerical algorithm is developed
allowing an estimation of IoU between two
arbitrarily oriented rectangles. Besides this, the algorithm is capable
of being implemented in GPU for rapid computation.

Recently,  \cite{li_rotation-insensitive_2017}
design multiangle anchor boxes to handle the problem of
rotation variation which is similar to ours. However, the anchor
boxes are limited to counteract orientation variation, and thus
orientation information is not predicted in the final output.
\cite{tianyu_tang_arbitrary-oriented_2017} take different framework (SSD) to obtain similar output for vehicle detection.
 In their work, calculation of IoU overlap between any two arbitrarily oriented rectangles  is technically avoided, which comes as our second contribution.

Beyond remote sensing, there is a similar problem in text detection where
text region in the given images are not always horizontal which
encourages researchers to work for more specific detection methods.
Concurrently with our work, \cite{ma_arbitrary-oriented_2018}
and \cite{jiang_r2cnn:_2017} propose to use oriented box for text detection
based on CNN.

\cite{ma_arbitrary-oriented_2018} 
present Rotation Region Proposal Networks (RRPN) to 
generate inclined proposals to include angle information. Following the 
two-stage fashion (e.g., Faster R-CNN\cite{ren_faster_2015}),
they develop Ration Region of Interest (RRoI) pooling  layer
to handle the arbitrary-oriented proposals generated by RRPN.
In their work, an algorithm for computing IoU between arbitrary-oriented
rectangles is also developed. Different from Faster R-CNN,
box re-regression has been cut off in their work leaving classifiers
for binary classification.
 
\cite{jiang_r2cnn:_2017} propose Rotational Region CNN (R2CNN) for
detecting text on natural images. In their work, anchor keeps axis-aligned 
when produced by RPN and orientation is taken into account by
imposing several pooling operation of different sizes to get
concatenated feature used for inclined box regression.

Both of the two works are based on Faster R-CNN and run a binary classification.
Similarly, our work also follows this fashion but takes a more
light and unified framework.
Our model inherits the anchor's generation but we truncate 
the pipeline to get one-stage detection model.
The model extends the predecessor's 4-dimension regression vector to 5-dimension
to incorporate orientation regression. In our work, all 
the 5 elements are simultaneously predicted when the convolution kernel
are sliding on the feature map as  we emphasize the ability of convolutional 
layers to locate object on the feature map which coincides with 
the research of \cite{Lenc_R-CNN_2015}. As a result, no further 
pooling is needed.

%
%

%% file: proposedMethod/proposedMethod.tex
\section{Proposed Method}\label{sec:proposed_method}
 \begin{figure*}
 	\centering
 	\includeImgPrefix{width=\textwidth}{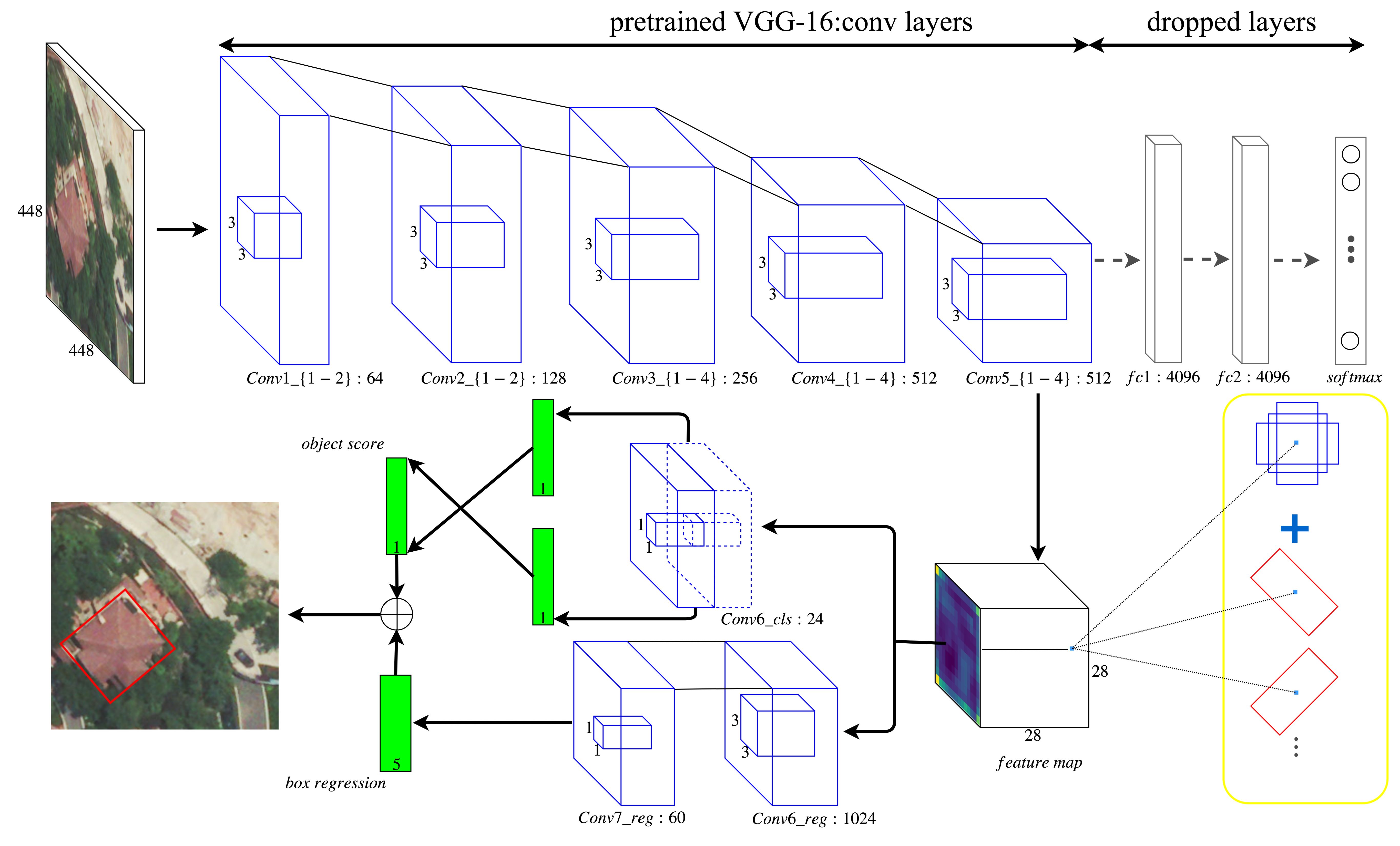}
 	\caption{Architecture of the proposed model. The pretrained VGG-16 model's 
 	parameters are used for feature extraction after being cut off its
 	classification layers. The extracted feature map goes through the two
 	extra convolutional layer branches (i.e., \textit{Conv6\_cls} for classification branch,
 	\textit{Conv6\_reg} and \textit{Conv7\_reg} for regression branch).
 	The feature map is shared between the both branches and our proposed multiangle anchors 
 	are generated at each pixel in the feature map.}
 	\label{fig:overview-model}
 \end{figure*}

Our proposed model is illustrated in \figurename{\ref{fig:overview-model}}.
For feature extraction, the pretrained VGG-16 model\cite{simonyan_very_2014}
is loaded after being cut off its fully connected layers and softmax layer.
The convolutional layers take an image of 448$\times$448
pixel size as input to produce a 512-channel 28$\times$28 pixel size feature map.
The following two extra convolutional layer branches keeping outputs' dimensionality 
same to that of the input are added to adapt the network to this task.
Network-based detection model usually consists of classification and
regression components. The feature map is to be shared
by classification (\textit{Conv6\_cls}) and regression 
(\textit{Conv6\_reg} and \textit{Conv7\_reg}) branches.
Each branch has its own layer configurations for its specific task.
In classification branch, the last convolutional layer's output
is split into two volumes, after which each volume is reshaped to a 
vector. A softmax is applied on the two to obtain the vector
of object score where each element is considered to measure 
the membership of the content in the corresponding bounding box
to the class of building object. The regression branch takes
one convolutional layer to assign a 5-dimension vector to 
each anchor box to get the bounding box. 
Each vector represents the predicted offset of the target building
from the already generated multiangle anchor which will be
introduced later.

\input{proposedMethod/subsec1}

\input{proposedMethod/subsec2}
\input{proposedMethod/subsec3}

%% file: proposedMethod/subsec1.tex
\subsection{Multiangle Anchor Box Proposal}

 \begin{figure}
 	\centering
 	\includeImgPrefix{width=0.45\textwidth}{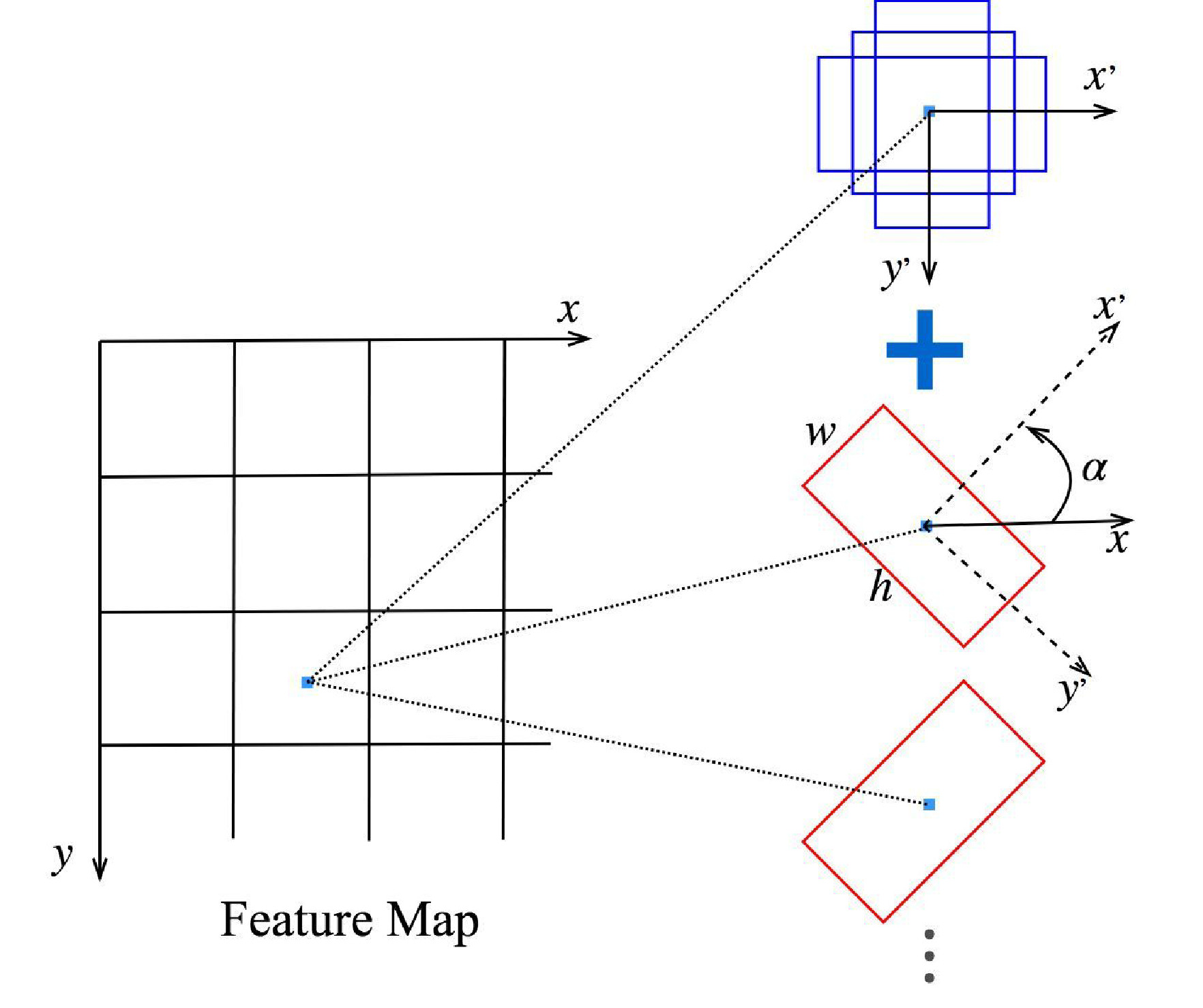}
 	\caption{Illustration of our proposed multiangle anchors. For
 		each location in the feature map we generate a fixed number of
 		anchors that consist of traditional ones (blue boxes) plus
 		oriented anchors (red boxes). The orientation angle is denoted
 		by $\alpha$.}
 	\label{fig:angleAnchor}
 \end{figure}
To incorporate angle information, we further develop RPN to
generate oriented anchor boxes in addition to the traditional ones.
 As \figurename{\ref{fig:angleAnchor}} illustrates, 
 on the shared feature map we generate our proposed 
 multiangle anchor boxes for each location.

The proposed boxes consist of axis-aligned and oriented
anchor boxes. 
They share a uniform format of $(x, y, \alpha, h, w)$, 
where $x$, $y$ are
measured by location of the center on the raw input
image; $\alpha$ is the angle between the
shorter side and the $x$ axis of the feature map; we denote the
longer side as $h$, subsequently the shorter one, $w$. To take account
of various sizes of buildings, we set different scales to configure
anchor boxes' shapes.
 To avoid ambiguity, we limit the value of orientation
angle in $[0, \pi)$.

%% file: proposedMethod/subsec2.tex
\subsection{IoU Estimation}\label{sec:fast-iou}
The value of IoU is frequently referenced in detection task.
Contrary to the axis-aligned boxes, calculating IoU for 
arbitrarily oriented boxes is to take longer time to be done and requires more
complex design in algorithm which may hamper the training and
inference process.

\begin{algorithm}
	\SetAlgoLined
	{\bf Input}: {$\bm{R_1}(x^r_1,y^r_1,\alpha^r_1,h^r_1,w^r_1)$,$\bm{R_2}(x^r_2,y^r_2,\alpha^r_2,h^r_2,w^r_2)$}\\
	{\bf Output}: IoU value\\
	{\bf Parameter}: $N$ //controls preciseness\\
	
	\begin{enumerate}
		\item[1:] 	Generate point vector $\bm{P}(\bm{P}_1 , \bm{P}_1 ,\cdots, \bm{P}_{N\times N} )$ in which $\bm{P}_i(x_i, y_i)$ is evenly 
		distributed in$	[−0.5,0.5]^2$
	\item[2:] 
	$\bm{P} \longleftarrow (x^r_1 , y^r_1 )^T + \bm{diag}(w^r_1 , h^r_1 )
	\bm{M}(\alpha^r_1)\bm{P}$ // rotate, scale and shift $\bm{P}$,  $\bm{M}$ denotes Rotation Matrix
	\item[3:] $(u , v ) := \bm{M}(-\alpha^r_2 )(x^r_2, y^r_2)^T$
	// rotate $-\alpha^r_2$ to get axis-aligned $\bm{R}_3$
	
	\item[4:] $\bm{P} \longleftarrow \bm{M}(-\alpha^r_2 )\bm{P}$
	// rotate same angle as $\bm{R}_2$
	
	\item[5:] $\bm{R}_3 := (u,v,0,h^r_2,w^r_2)$
	//after this operation $\bm{R}_3$ 	is axis-aligned
	\item[6:] $n$: the number of points in 
	$\bm{P}$ which fall into $\bm{R}_3$ 
	\item[7:] $I := h^r_1w^r_1n/N^2$// calculate the intersection area
	
	\item[8:] $IoU:=I/(h^r_1w^r_1+h^r_2w^r_2-I)$
	\item[9:]\textit{\bf return} $IoU$ 
	\end{enumerate}	

	\caption{IoU Estimation}
	\label{algo:algorithm1}
\end{algorithm}

 \begin{figure} 
 	\centering
 	\includeImgPrefix{width=0.5\textwidth}{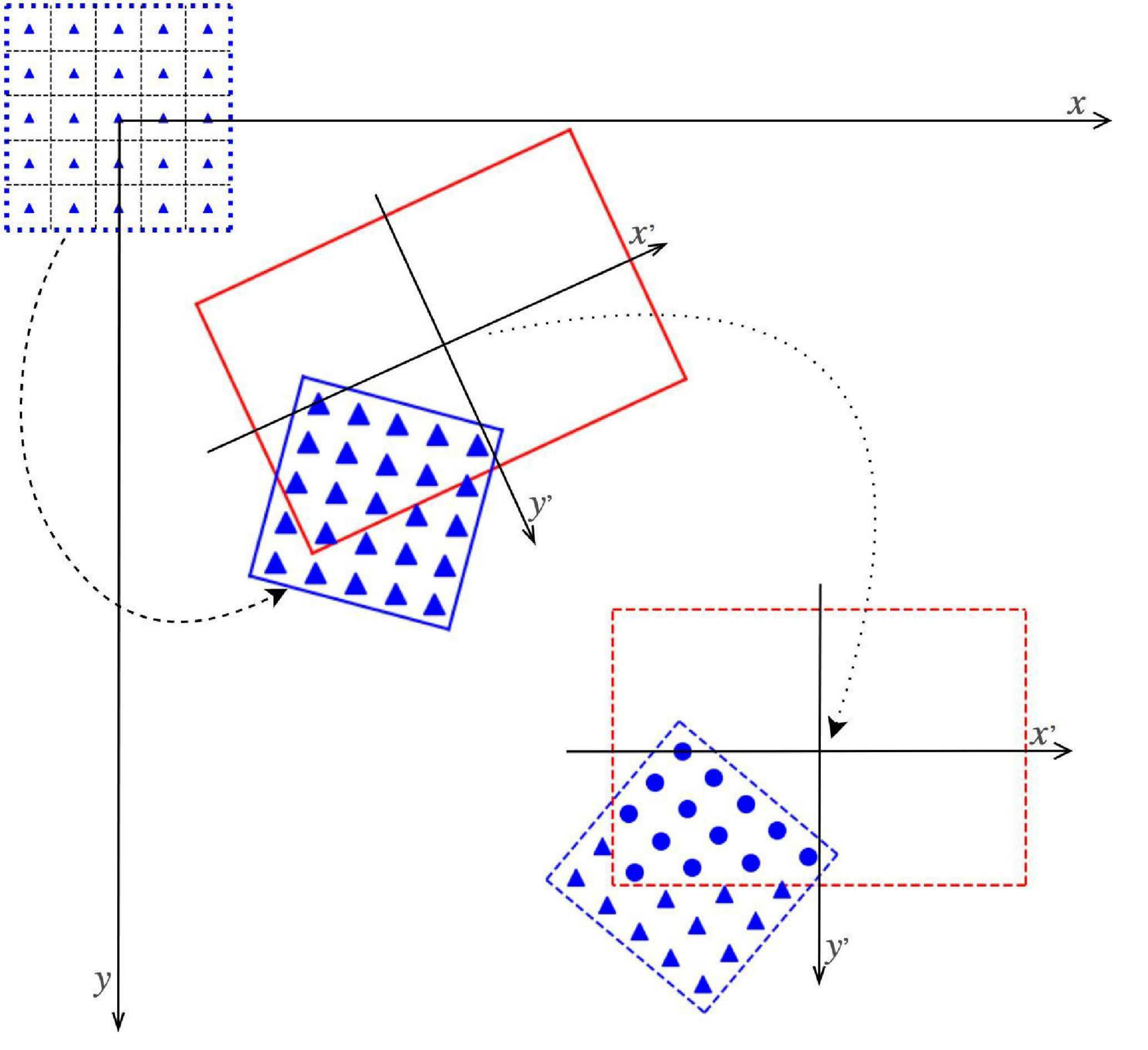}
 	\caption{Illustration of our fast estimation algorithm. To estimate
 		two arbitrarily imposed rectangles (denoted by red and blue solid
 		boxes respectively, center of the figure) we firstly generate grid
 		points (blue triangles, up-left) and transfer them to fit one
 		(triangles, center). After another transformation (down-right) we
 		count the intersection points (circles).}
 	\label{fig:fastIoU}
 \end{figure}

To tackle this problem, we develop a numerical algorithm to
estimate IoU between two arbitrarily oriented rectangles
and the precision could be controlled via a single parameter,
named \textit{Fast-IoU}. The algorithm is illustrated in  Algorithm\ref{algo:algorithm1} and \figurename{\ref{fig:fastIoU}}.
 The ideas behind the algorithm are using point
number to approximate areas and simplifying counting inner
points by rotation transformation. As showed in \figurename{\ref{fig:fastIoU}}, in the
first stage, a group of uniformly distributed gird points are
generated in a unit square which are subsequently transferred to
fit the inner space of one of the two rectangles by an affine matrix.
By applying another affine transformation on the inner points and
the other box, we make the box axis-aligned while preserving the
whole geometry property, after which counting the inner points
of the box becomes straightforward.
 \begin{figure*}
 	\centering
 	\centering
 	\includeImgPrefix{scale=0.08}{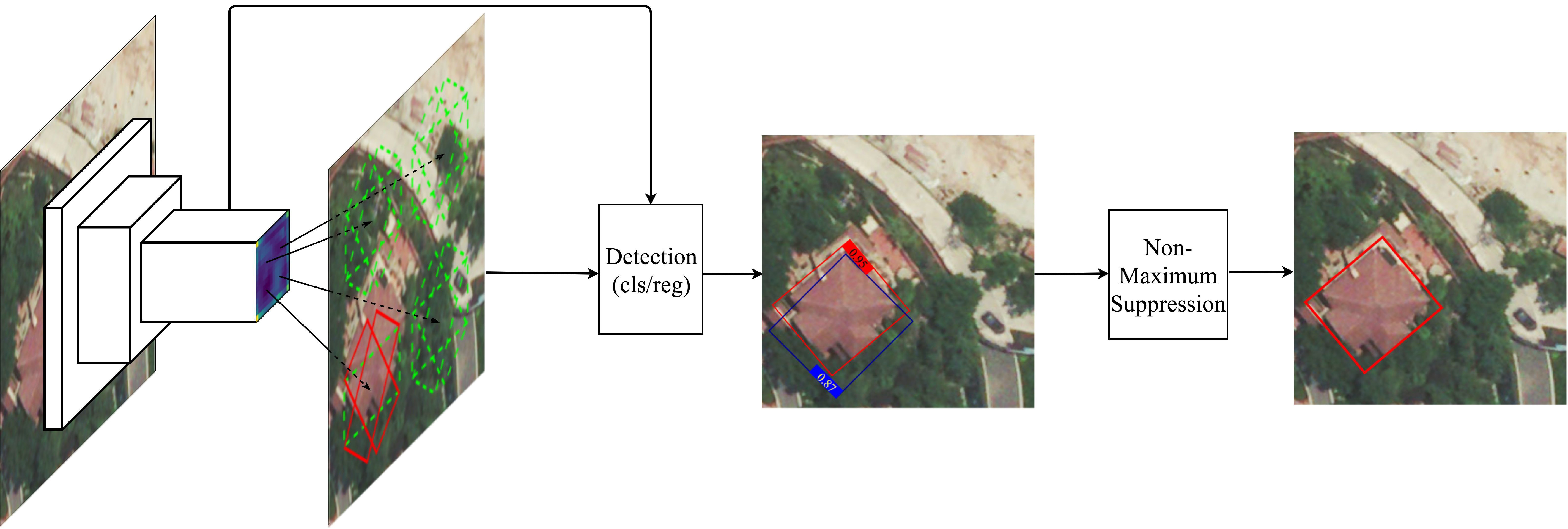}
 	\caption{Pipeline of the proposed method. For a given image, the feature map is obtained through a stack of convolutional layers and
 		oriented anchors are generated for each location of the feature map. Anchors with IOU below the threshold are marked as negative
 		samples (green-dashed boxes). The detection component predicts both class and offset for each anchor. For post processing \textit{non-maximum suppression} is applied.}
 	\label{fig:pipeline}
 \end{figure*}

With the multiangle anchor mechanism we further illustrate our method in \figurename{\ref{fig:pipeline}}. The method generates multiangle anchors on 
the feature map and uses the \textit{Fast-IoU}
to implement \textit{NMS} operation.

%% file: proposedMethod/subsec3.tex
\subsection{Training and Inference}
To train the proposed model we use loss function defined in 
\cite{girshick_fast_2015}:
\begin{eqnarray}
L(\{p_i\},\{t_i\}) = \frac{1}{N_{cls}}\sum_{i}^{}L_{cls}(p_i,l_i)
\nonumber\\
+ \lambda\frac{1}{N_{reg}}\sum_{i}^{}p^*_i
L_{reg}(t_i,t^*_i)
\end{eqnarray}
where $i$ denotes the index of anchor in training batch and, $p_i$ , $t_i$
represent predicted probability of the anchor being a building object
 and 5-dimension vector parameterizing geometry properties of the
predicted box respectively. $l_i$ and $t^*_i$ are the corresponding
ground truths.

Cross-entropy is used to evaluate classification loss
$L_{cls}(p_i,l_i)$:
\begin{eqnarray}
L_{cls}(p_i,l_i) = - \log{p_i^{(l_i)}}
\end{eqnarray}
and for regression loss, we use $smooth~L_1$:
\begin{eqnarray}
L_{reg}(t_i, t^*_i) = \sum_{k\in\{x,y,\alpha,h,w\}}^{}w_kL_1(t^{(k)}_i-
t_i^{*(k)})\\
L_1(x)=\begin{cases}
0.5x^2&if~|x|<1,\\
|x|-0.5&else
\end{cases}
\end{eqnarray}
where, $w_k$ is the weight of the $k$ th dimension. Each dimension is
defined as:
\begin{eqnarray}
\begin{split}
t_x=\frac{x-x_a}{w_a} &,& t_y=\frac{y-y_a}{h_a}\\
t_h=\log{\frac{h}{h_a}} &,& t_w=\log{\frac{w}{w_a}},\\
t_\alpha=\alpha-\alpha_a &,& t^*_x=\frac{x^*-x_a}{w_a},\\
t^*_y=\frac{y^*-y_a}{h_a} &,& t^*_h=\log{\frac{h^*}{h_a}},\\
t^*_w=\log{\frac{w^*}{w_a}} &,& t^*_\alpha=\alpha^*-\alpha_a
\end{split}
\end{eqnarray}
where subscript of $a$ and
superscript of $*$ indicate anchor and ground-truth respectively.

We notice that each dimension is parameterized only by 
local information, independent from image size, which is
essential for adapting the trained model to larger input.

We exploit the strategy in \cite{ren_faster_2015}
 to assign labels to anchors. That
is, anchors with highest IoU overlap with any ground-truth box or
with IoU exceeding specific threshold are marked as targets.
Also, since buildings are sparse in an input image we take
sampling strategy from \cite{girshick_fast_2015}
 to generate training batch.
  With the network architecture depicted in  \figurename{\ref{fig:overview-model}},
  our training processing
 makes the model directly learn to regress oriented
 bounding box.

Contrary to the huge amount of training samples required for
training \textit{VGG-16}, in this task, we are only able to obtain a limited
number of manually labeled images. As a result, we take the fine
tuning strategy to facilitate the training process by loading the
convolutional layers' parameters of VGG-16.

In inference stage, we map the outputs (values of the offsets)
back to the original image to get rotated bounding boxes and
apply NMS for eliminating redundant boxes.
 The \textit{Fast-IoU}, again,
 is invoked to calculate IoU among the boxes.
  The
process greedily filters out boxes whose IoUs with their buddies
excess the predefined threshold while being associated with
lower predicted scores.

%% file: experiments/experiments.tex
\section{Experimental Results and Discussion}\label{sec:experiments}
In this section, we train our proposed model and run the
prediction on the test images. 
We compare the popular axis-aligned
 method with ours in both subjective and objective ways.
Though the experiments are conducted on a relatively small dataset
, the
results explicitly show that the proposed method outperforms its
counterpart
 in handling optical remote sensing images,
 more specifically, we use faster R-CNN\cite{ren_faster_2015}
 as the baseline method.
  We also
conduct experiments on 896$\times$896 images and the results show
that our model is able to learn to
 detect building object by local context information and is capable of
scaling to variable sizes of input images.

We also conduct experiment for R2CNN\cite{jiang_r2cnn:_2017} which is proposed
to address arbitrary-orientated text detection on our building dataset.
Results show an approximately equal performance of them two.

\input{experiments/subsec1}

\input{experiments/subsec2}
\input{experiments/subsec3}

\input{experiments/subsec4}

\input{experiments/subsec5}

%% file: experiments/subsec1.tex
\subsection{Dataset Setup}
 \begin{figure*}
 	\centering 	
	\datasetsubfig{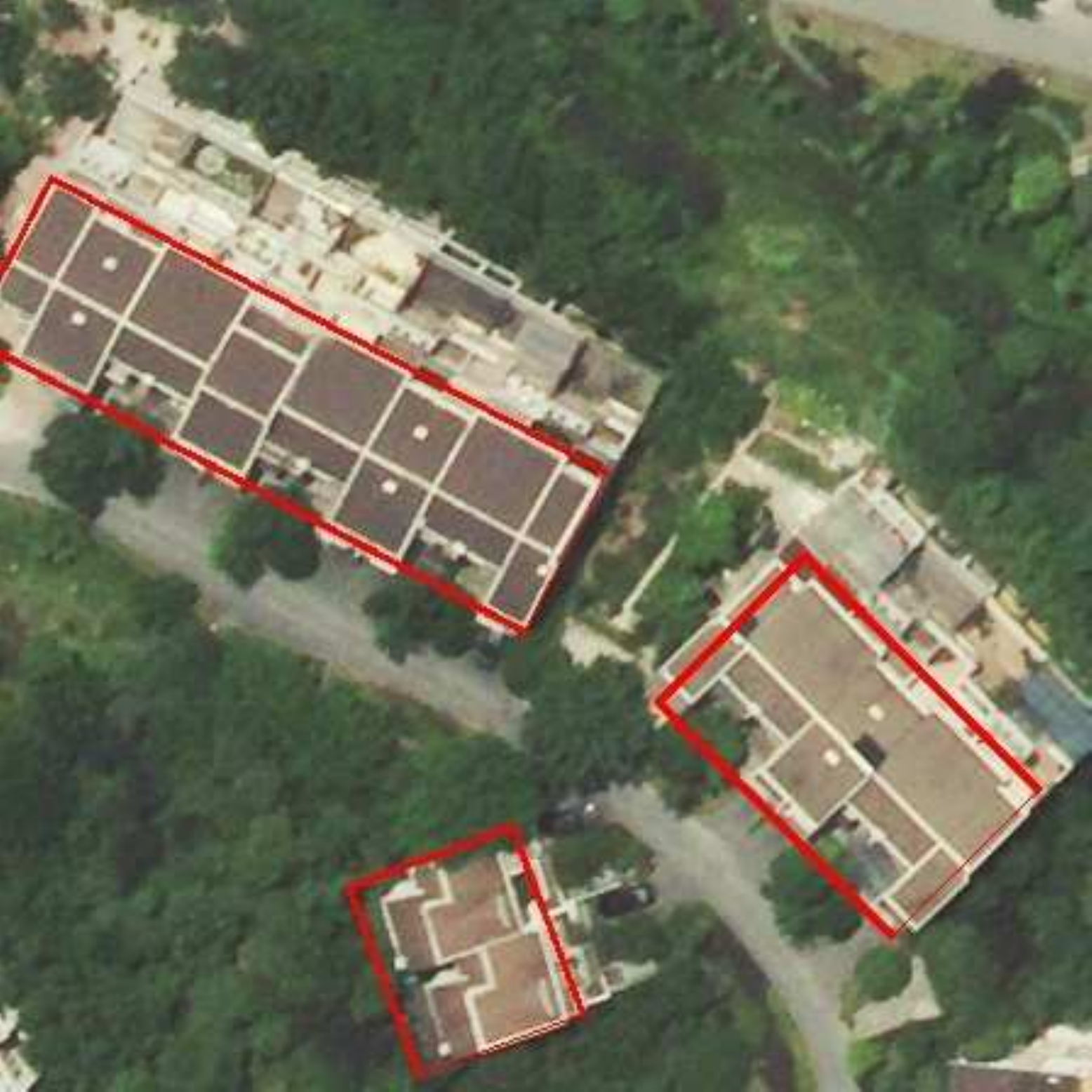}
	\datasetsubfig{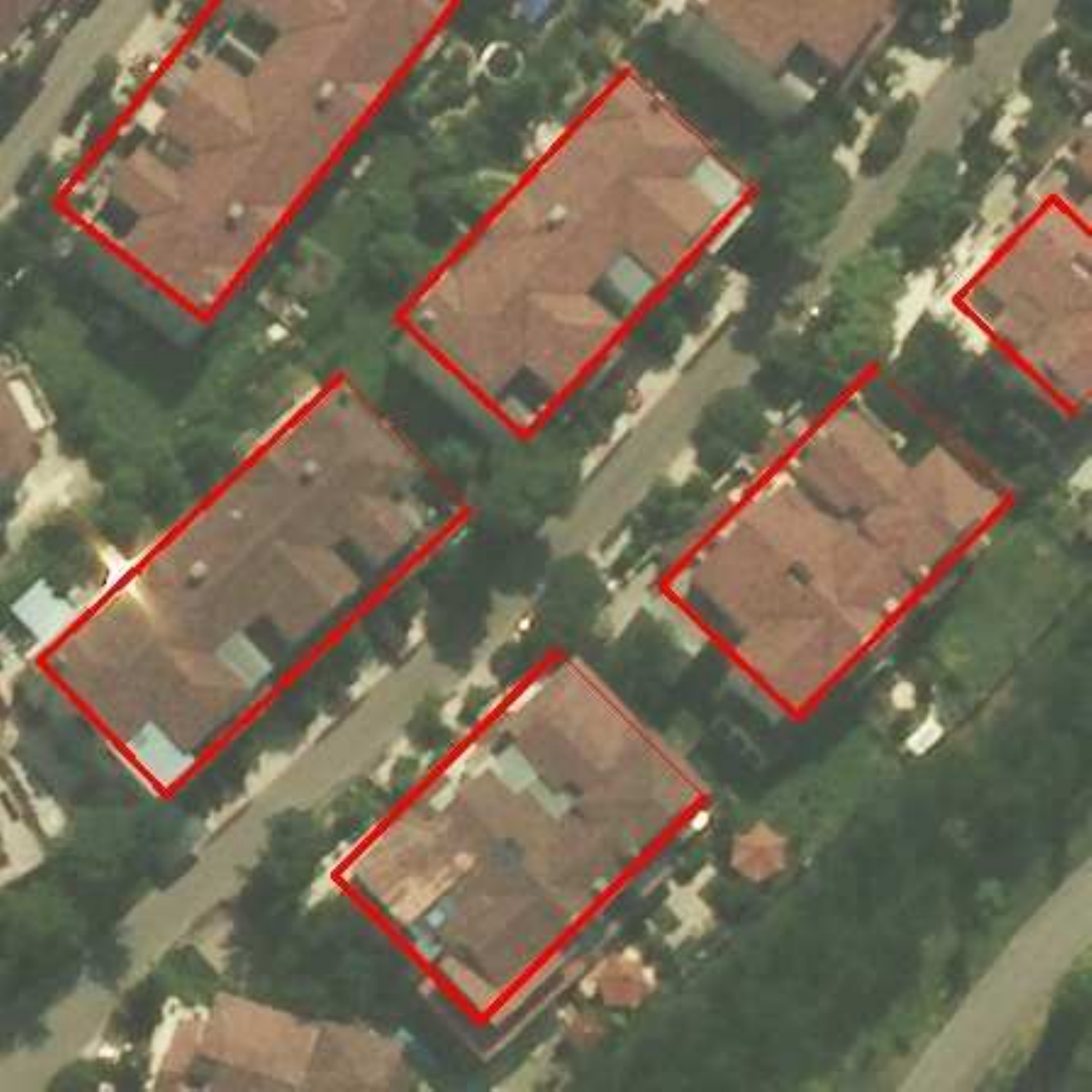}
	\datasetsubfig{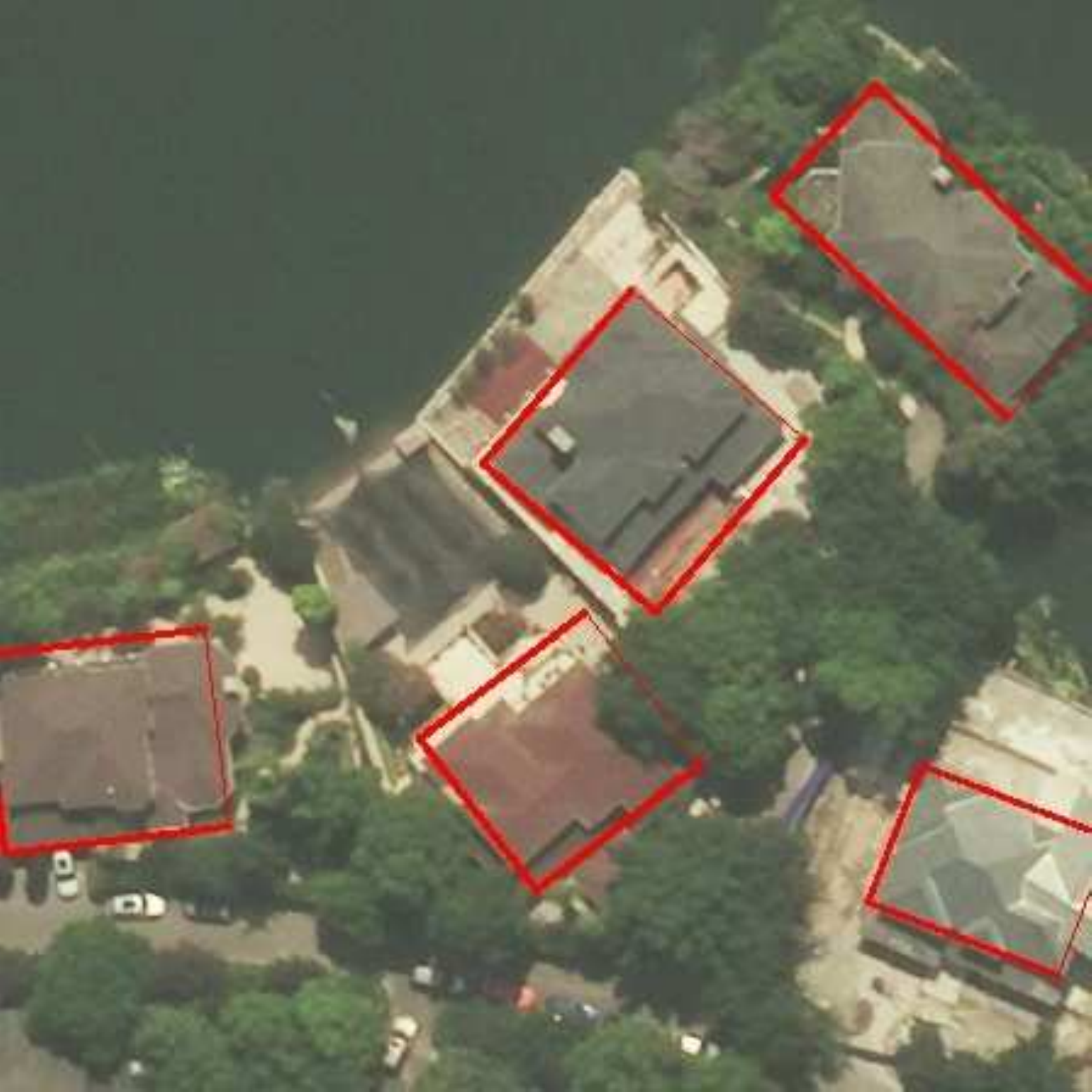}
	\datasetsubfig{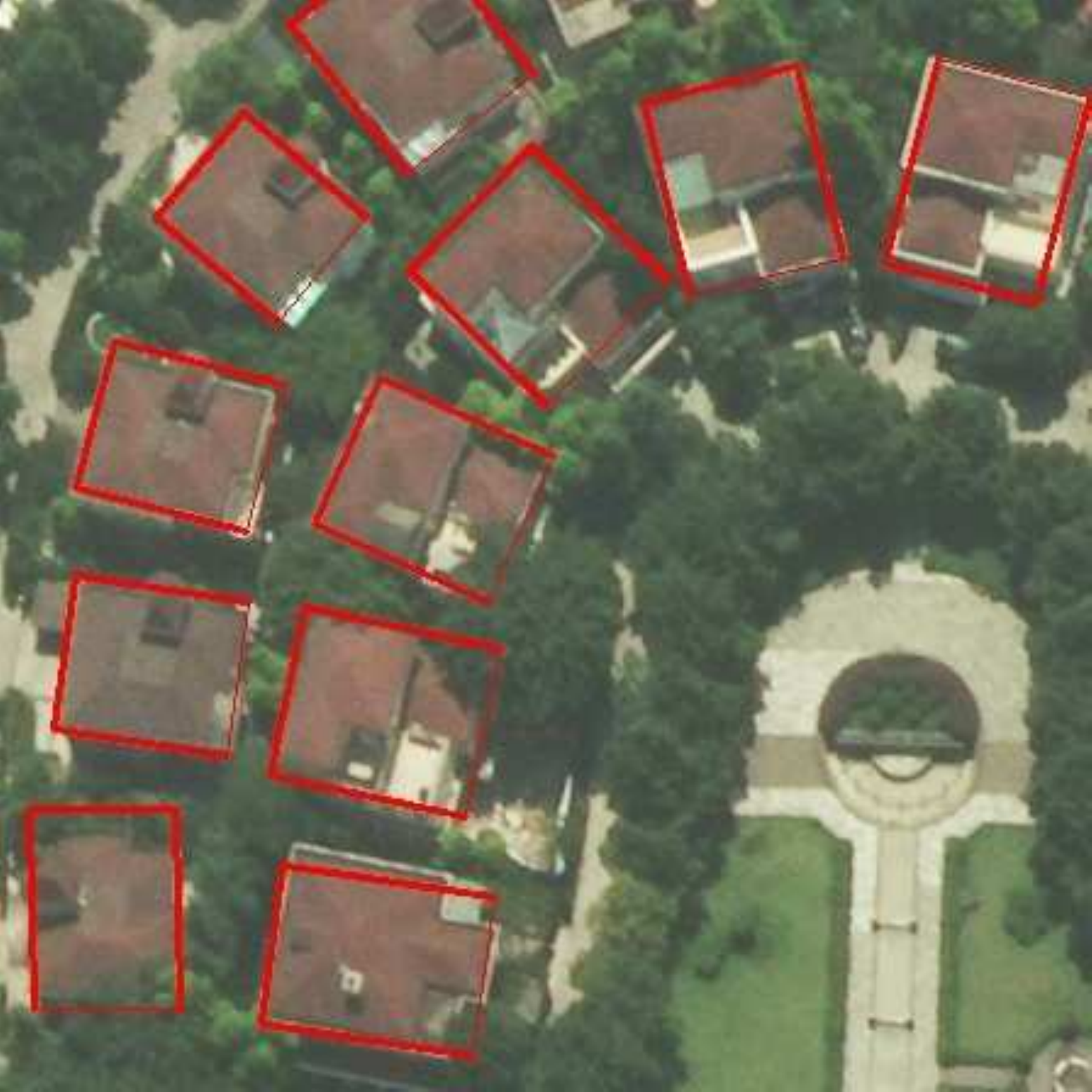}
	\datasetsubfig{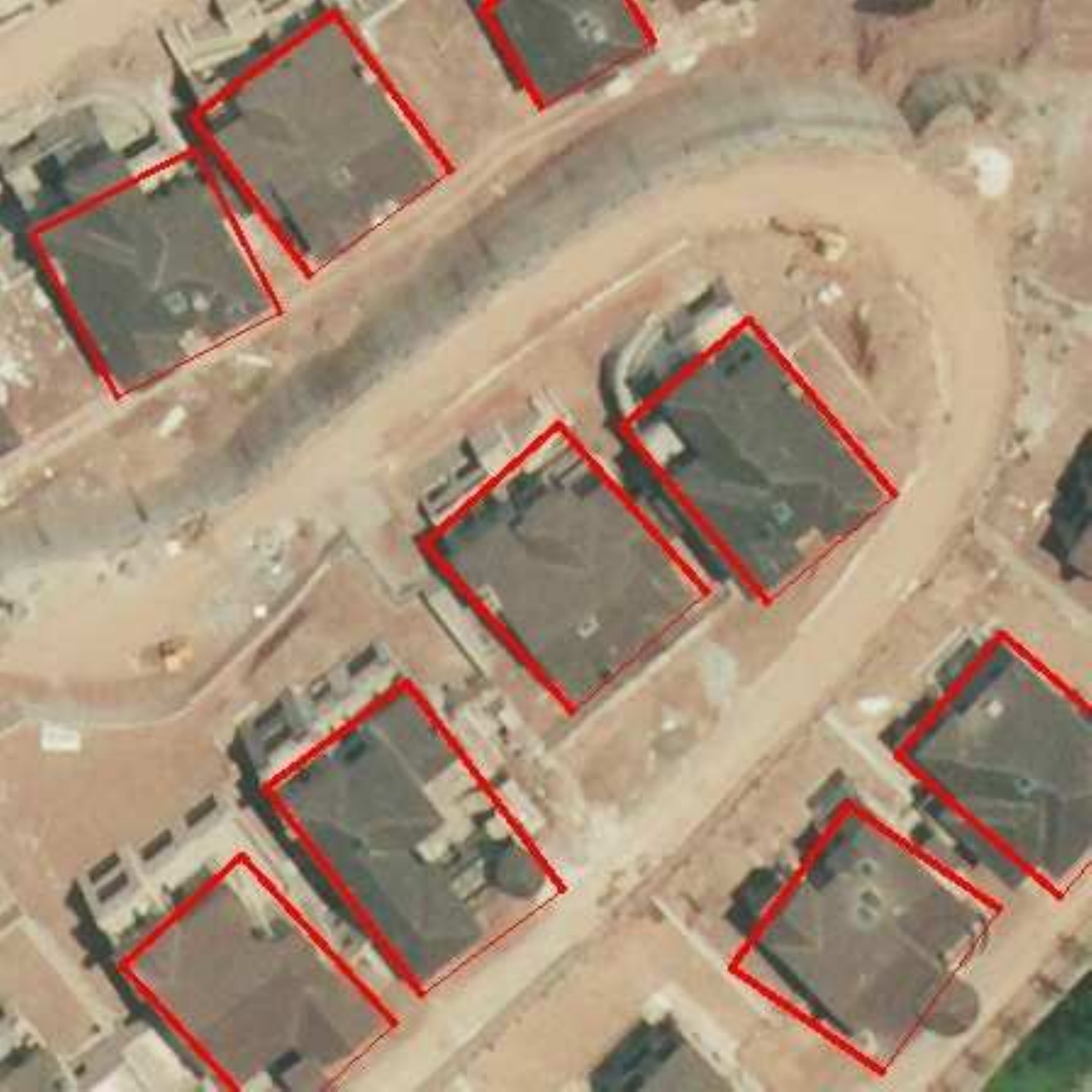}
\datasetsubfig{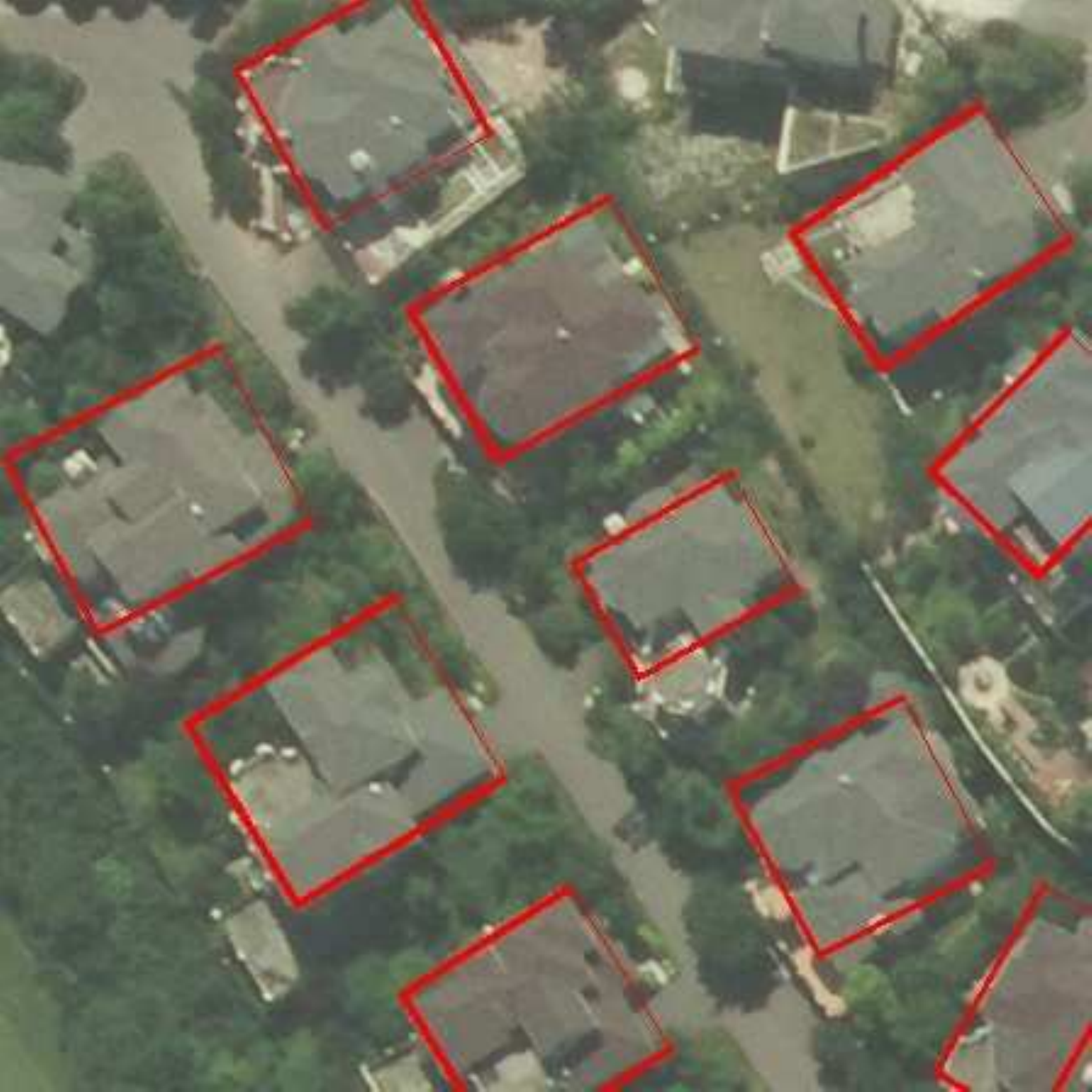}
\datasetsubfig{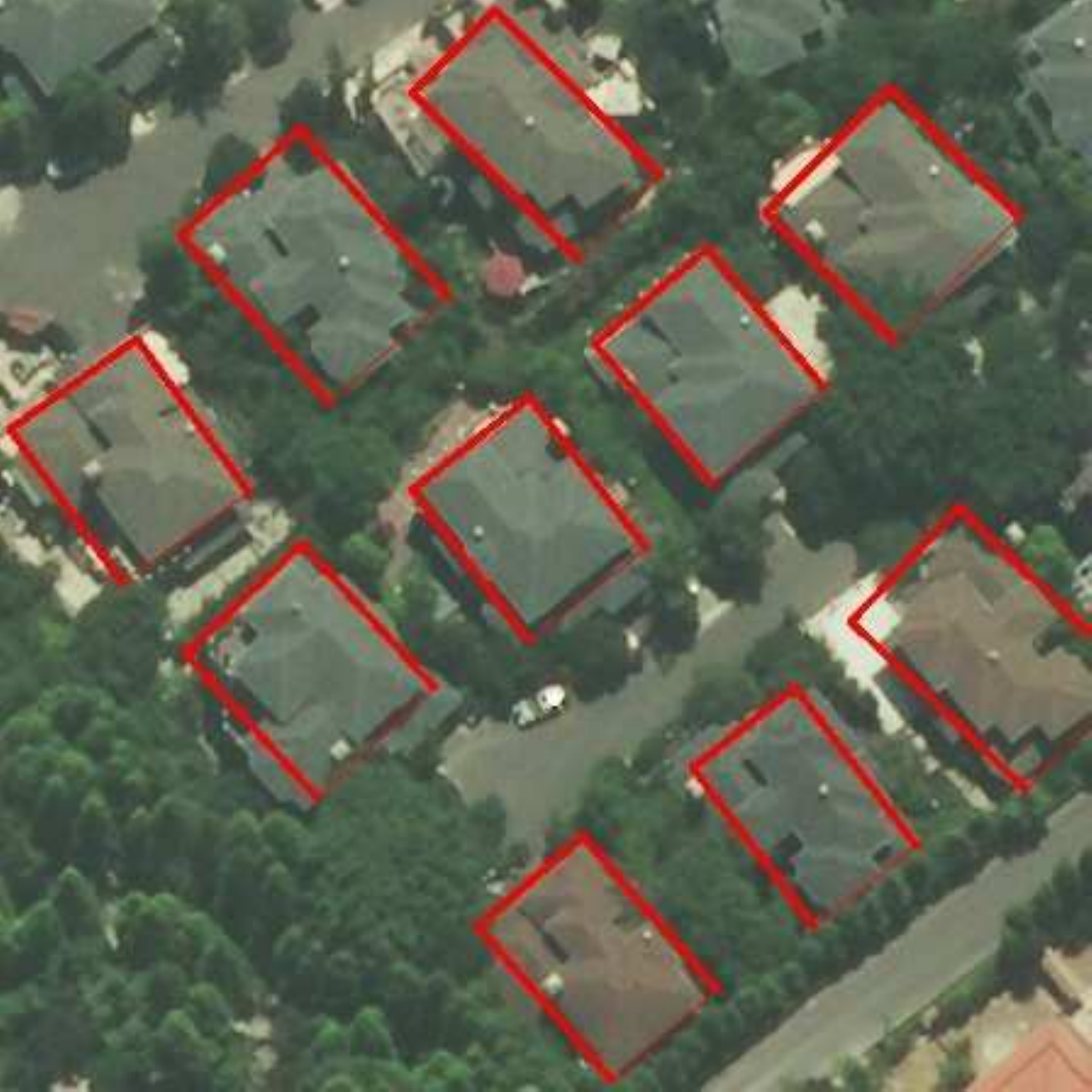}
 	 \\
\datasetsubfig{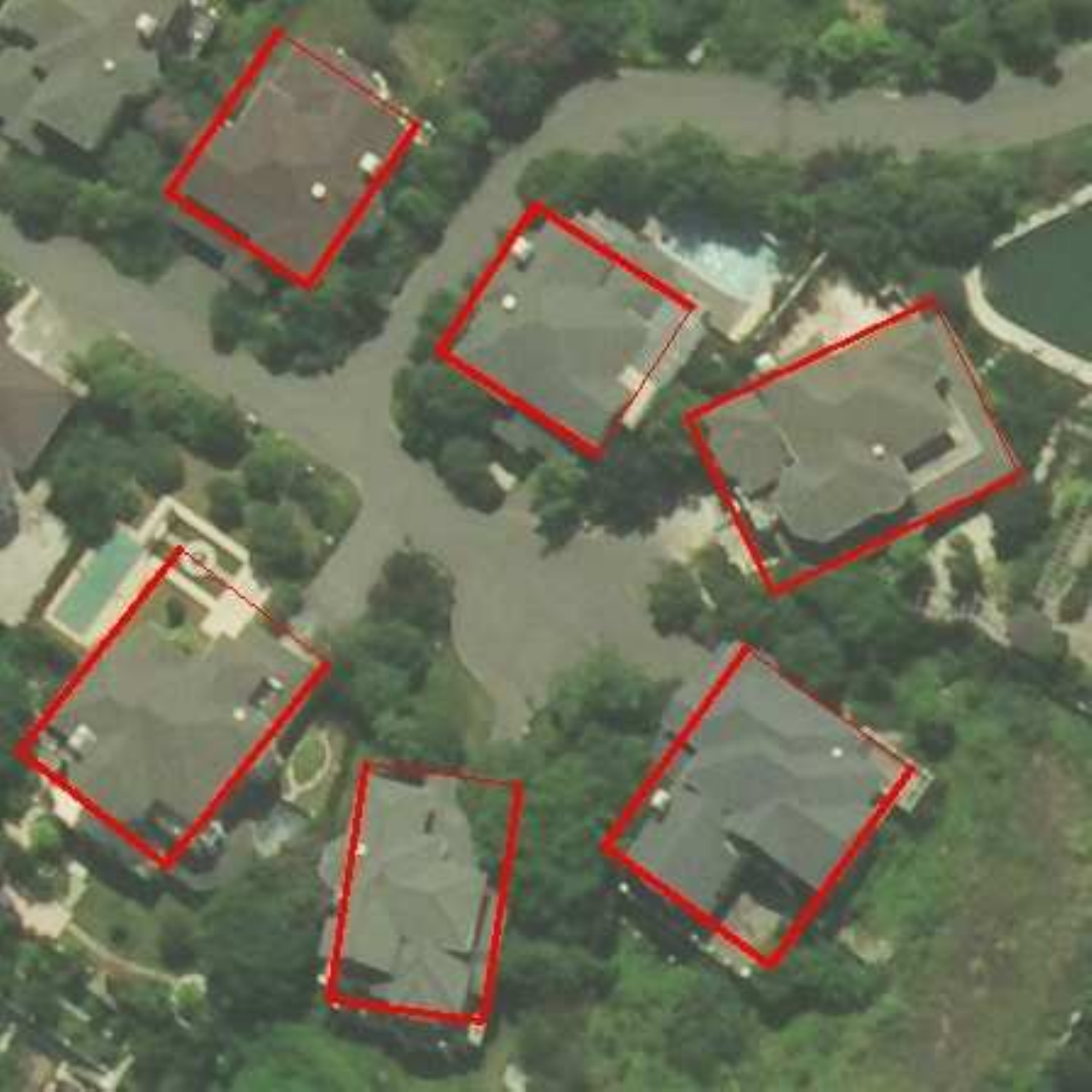}
\datasetsubfig{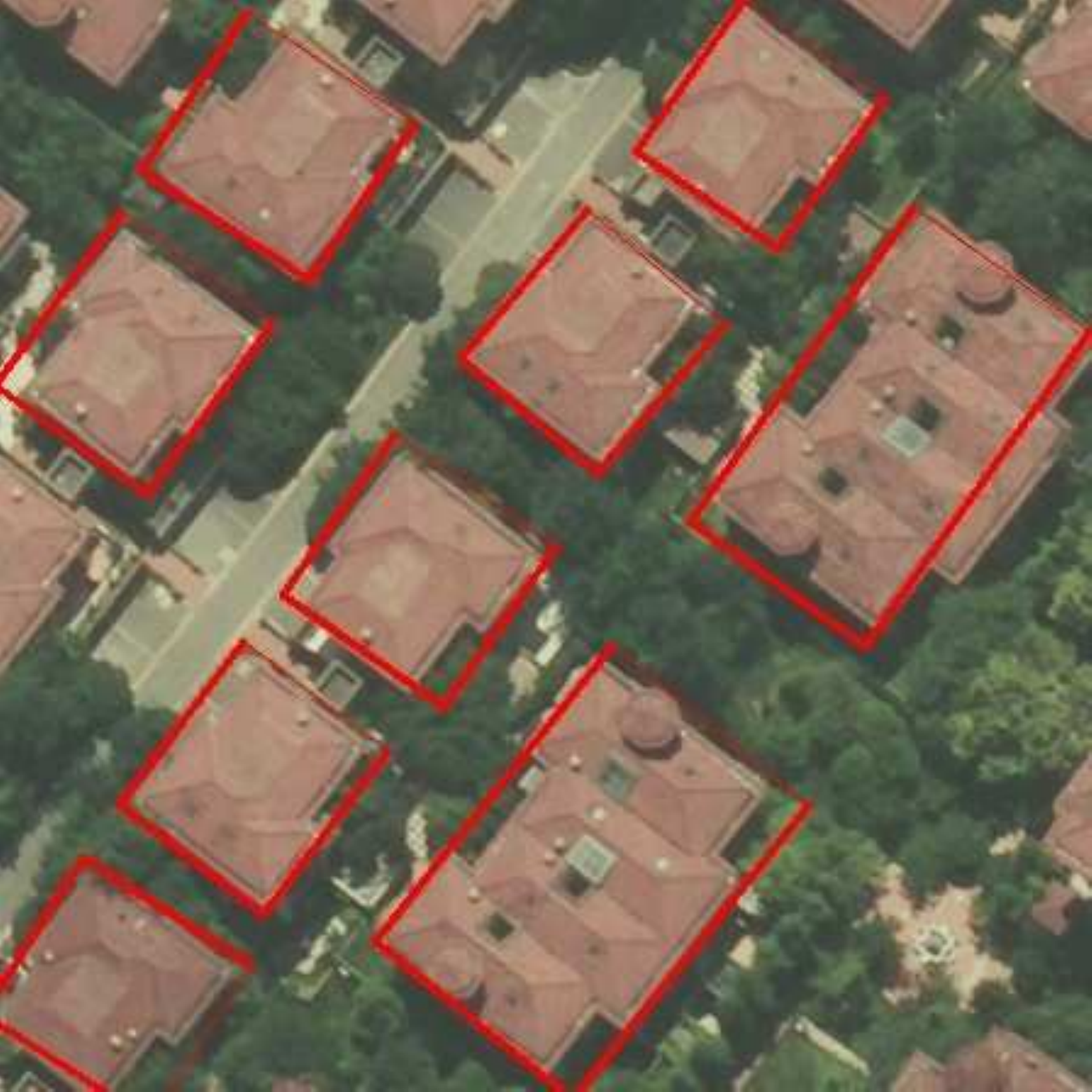}
\datasetsubfig{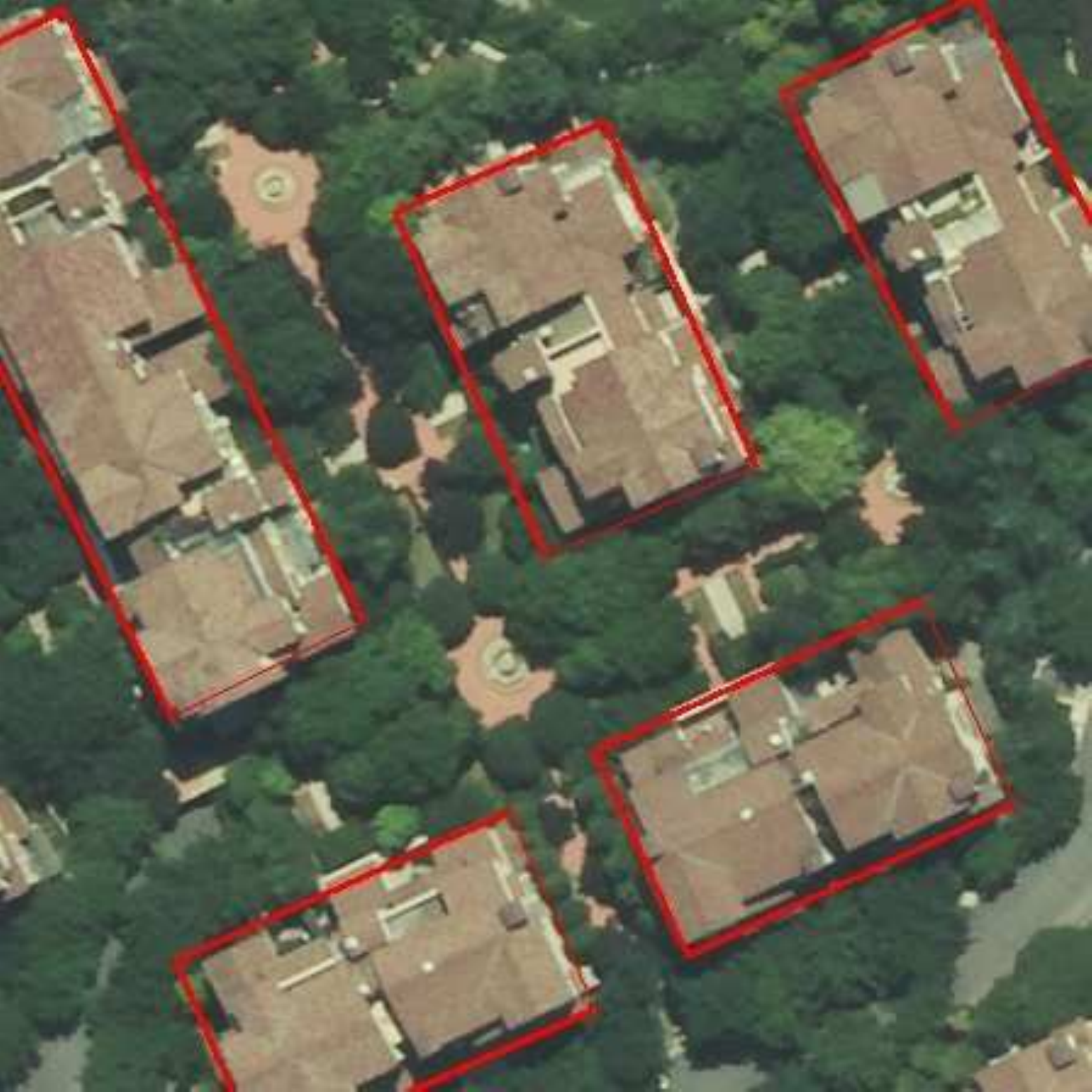}
\datasetsubfig{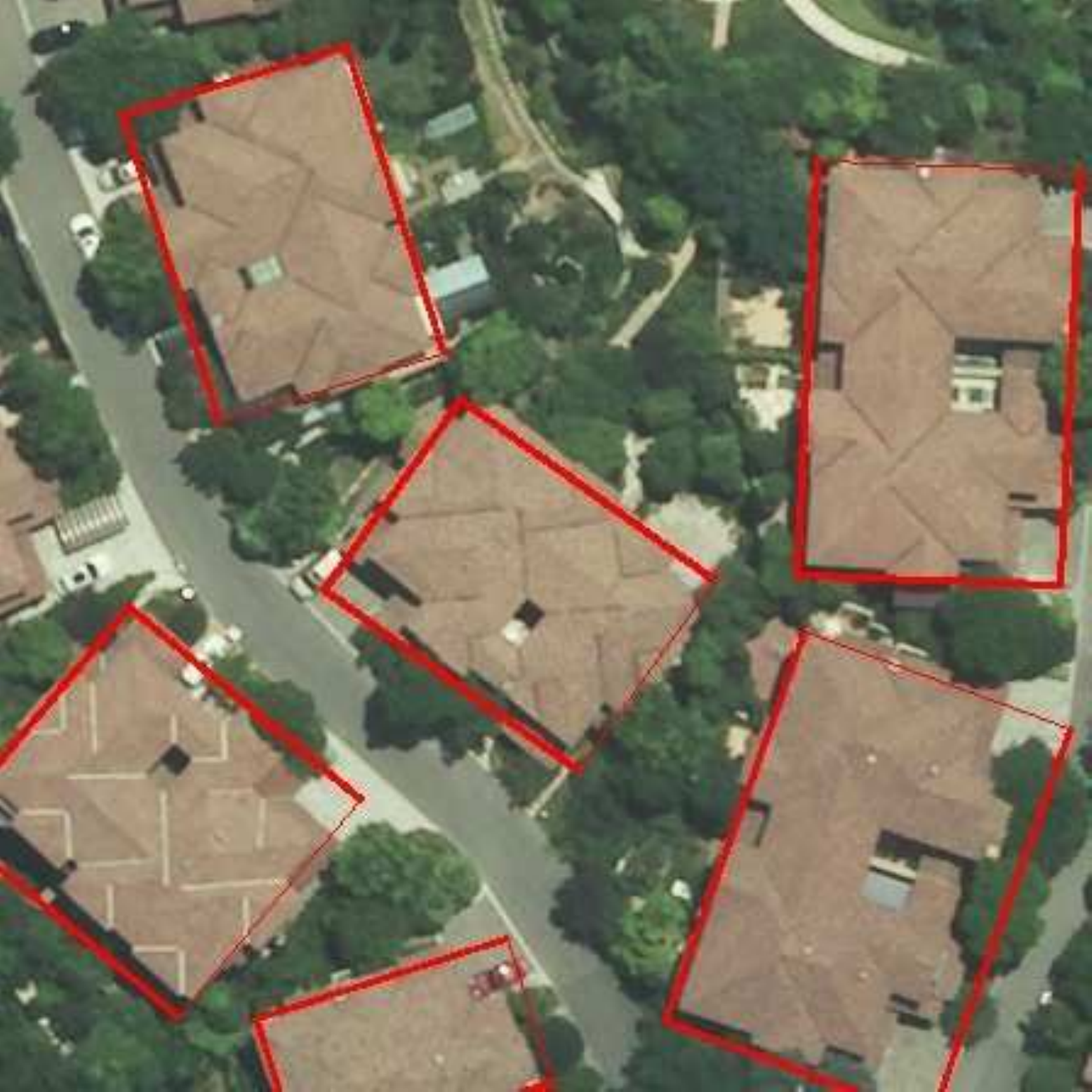}
\datasetsubfig{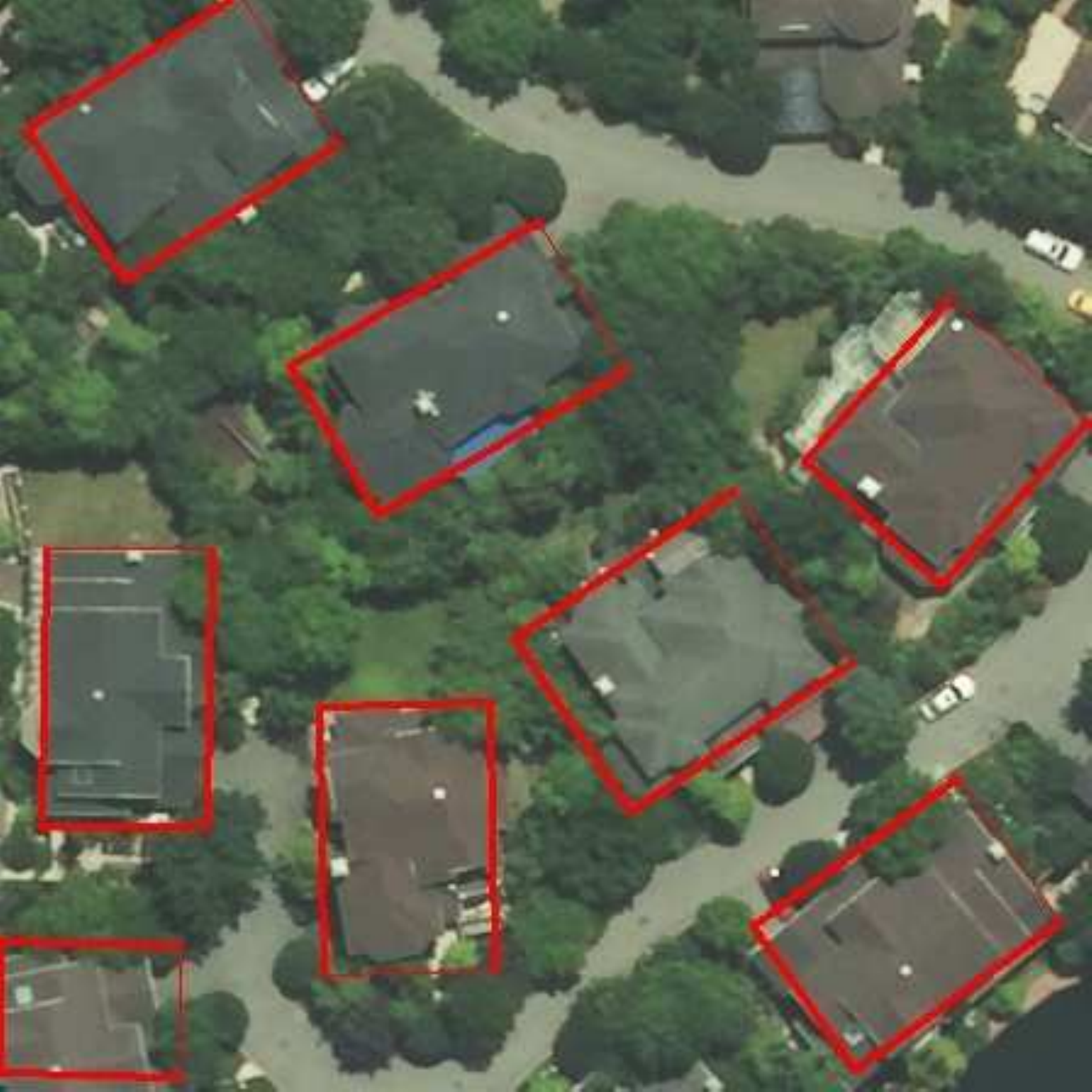}
\datasetsubfig{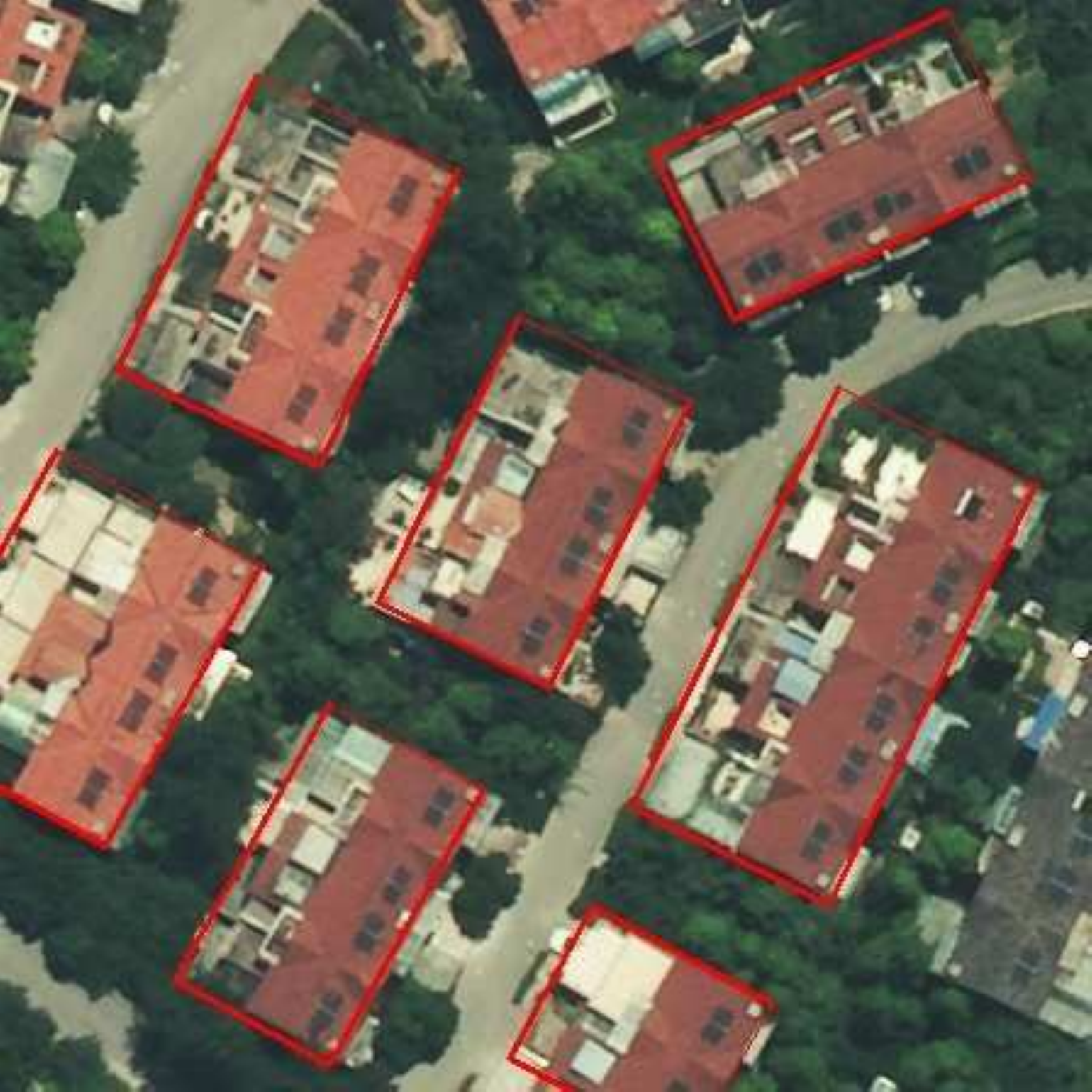}
\datasetsubfig{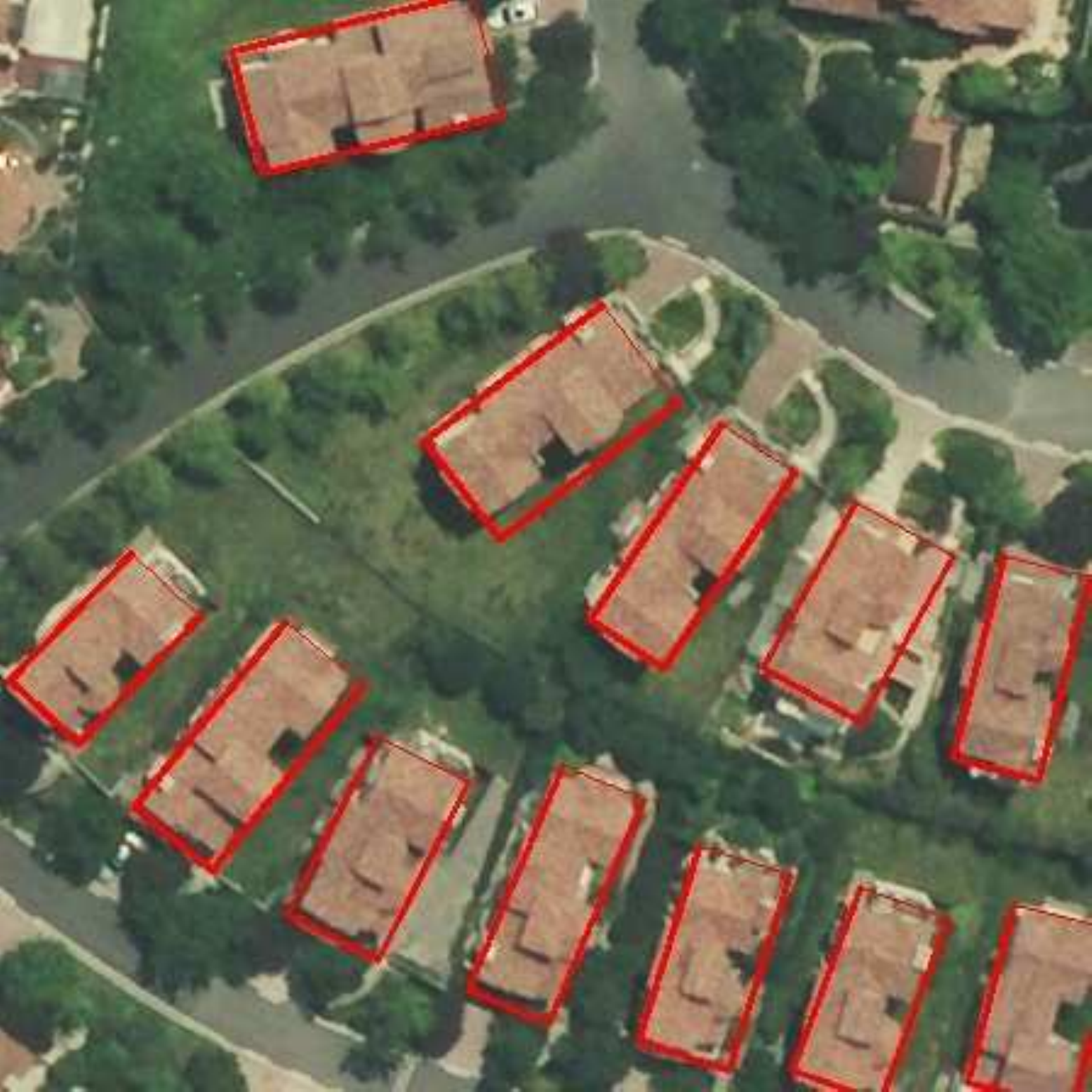}
 	\caption{ Part of labeled samples for training and evaluating.
 	 Each image has 448$\times$448 pixels, and target buildings are marked	by red bounding-boxes. 
 		Beyond the popular four-dimension, rotation angle is taken into account to locate target building.	}
 	\label{fig:sample-images}
 	 \end{figure*}

There is a scarcity of off-the-shelf datasets 
 for training and evaluating for this task, which requires orientation labels.
  To handle this
problem, we manually labeled 364 images of 448$\times$448 size,
target buildings in each image are enclosed within a rotated
bounding box. Some labeled samples are as shown in
 \figurename{\ref{fig:sample-images}}.
Each target building is described by a 5-dimension vector:
$(x, y, \alpha, h, w)$ where, $x, y$ denote the coordinate of the center,
 and $\alpha$ denotes the angle from
image's $x$-axis to bonding-box's $x$-axis.

%% file: experiments/subsec2.tex
\subsection{Implementation Details and Parameter Optimization}
Within \textit{back-propagation} our model can be trained in 
an end-to-end fashion and
we take stochastic gradient descent (SGD) for optimization. In
each updating step, the model takes one image as the input,
 all anchors are
marked as positive, negative or ignored (anchors with ignored
label make no effect in model's updating) and we randomly choose
128 anchors to evaluate the loss function.
 Since targets are sparse in real images we restrict the
ratio of the positive and negative samples to 1:1 to avoid overfitting.

For leveraging the CNN's power in feature extraction, we use
convolutional layers and the corresponding parameters of the
\textit{VGG-16} model pre-trained on ImageNet. The regression and
classification branches are shallow and randomly initialized for
training. However, for the borrowed convolutional layers, there
are some noticeable challenges in exploiting the pre-trained
parameters. Firstly, there is a considerable difference between
ordinary image and remote sensing image: objects in the two have
totally different orientation styles which makes feature extraction less
effective and thus hurts the performance in both classification
and regression; then, as we know, CNN is usually trained for
sight rotation invariance which is achieved by installing pooling
layers and mirroring training images. However, in this task we aim
to output a rotation angle for each target which moves beyond
that property. Those encourage us to keep these layers
learnable with slowing their learning rates to 1/100 of their following
branches' (i.e., classification and regression).

%% file: experiments/subsec3.tex
\subsection{Evaluation Metrics}
To evaluate the performance of the proposed method, two
widely-used measures are used, namely the precision-recall
curve (PRC) and average precision (AP).

For \textit{Precision-Recall Curve}, we use \textit{TP}, \textit{FP} and
 \textit{FN} to denote
the number of true-positives, false-positives and false-negatives
respectively. Precision measures the fraction of true-positives
(TPs) over all samples that are predicted as positives:
\textit{Precision=TP/(TP+FP)}.
 And Recall measures the fraction of TPs
over all sample that are labeled as positives:
\textit{Recall=TP/(TP+FN)}.
The AP measures the average value of 
precision over the interval ranging from 
\textit{Recall=0} to \textit{Recall=1}.
Hence, higher curve for PRC and higher value for AP are
desired.

Generally, under the topic of object detection, a detection
box whose IoU overlap ratio with any ground truth is greater
than 0.5 is considered to be a true positive, and otherwise false positives.
For several boxes overlapping with the
same ground truth we only consider the one of the highest
overlap ratio as a true positive.

%% file: experiments/subsec4.tex
\subsection{Experiments and Discussions}
In this subsection, we aim to demonstrate that with the help of
rotated anchor mechanism, our proposed method tends to be
more efficient in locating object over its axis-aligned 
counterpart. The IOU value is calculated by the algorithm
proposed in subsection \ref{sec:fast-iou}.

\begin{figure} 
\centering 	
\expfirstcmp{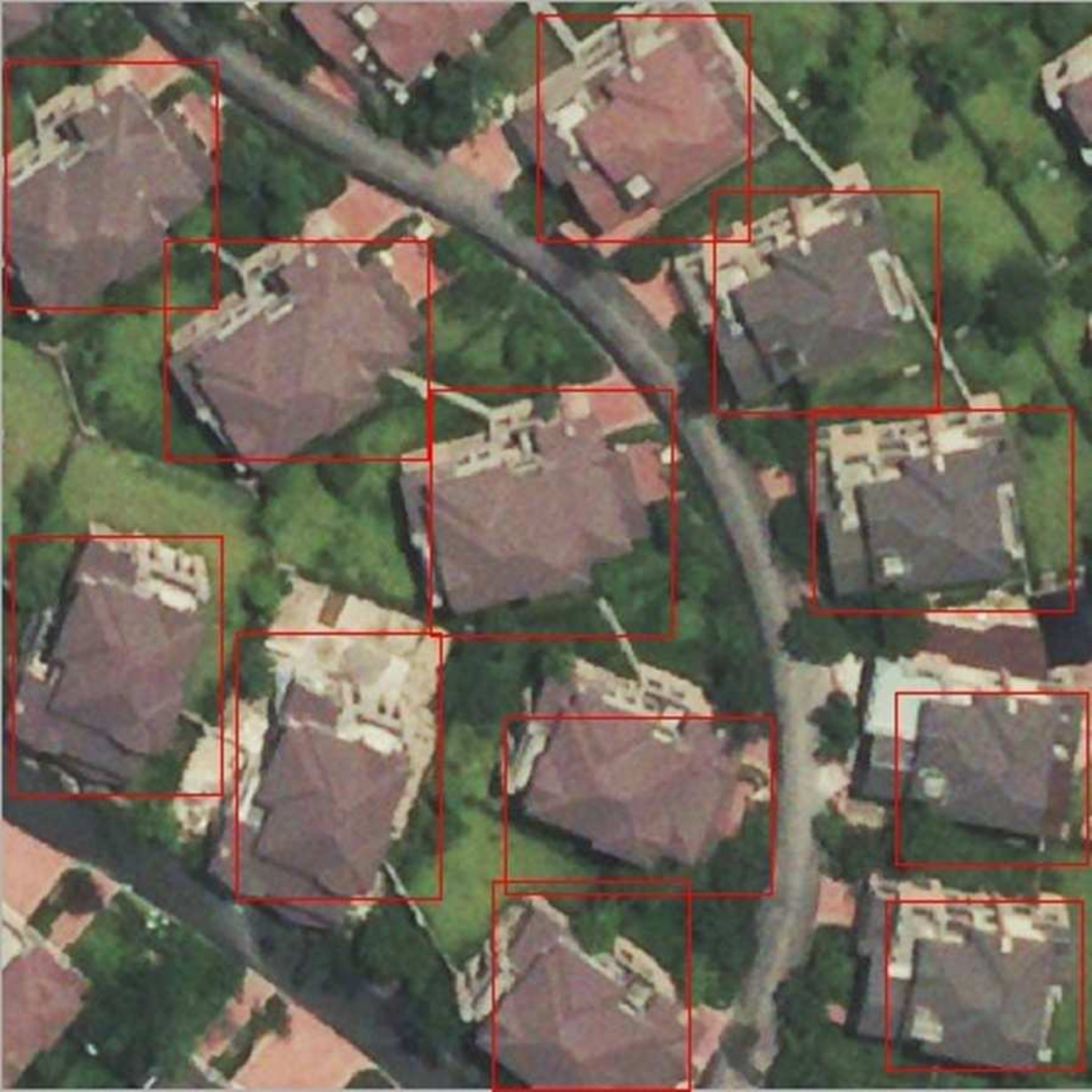}
\expfirstcmp{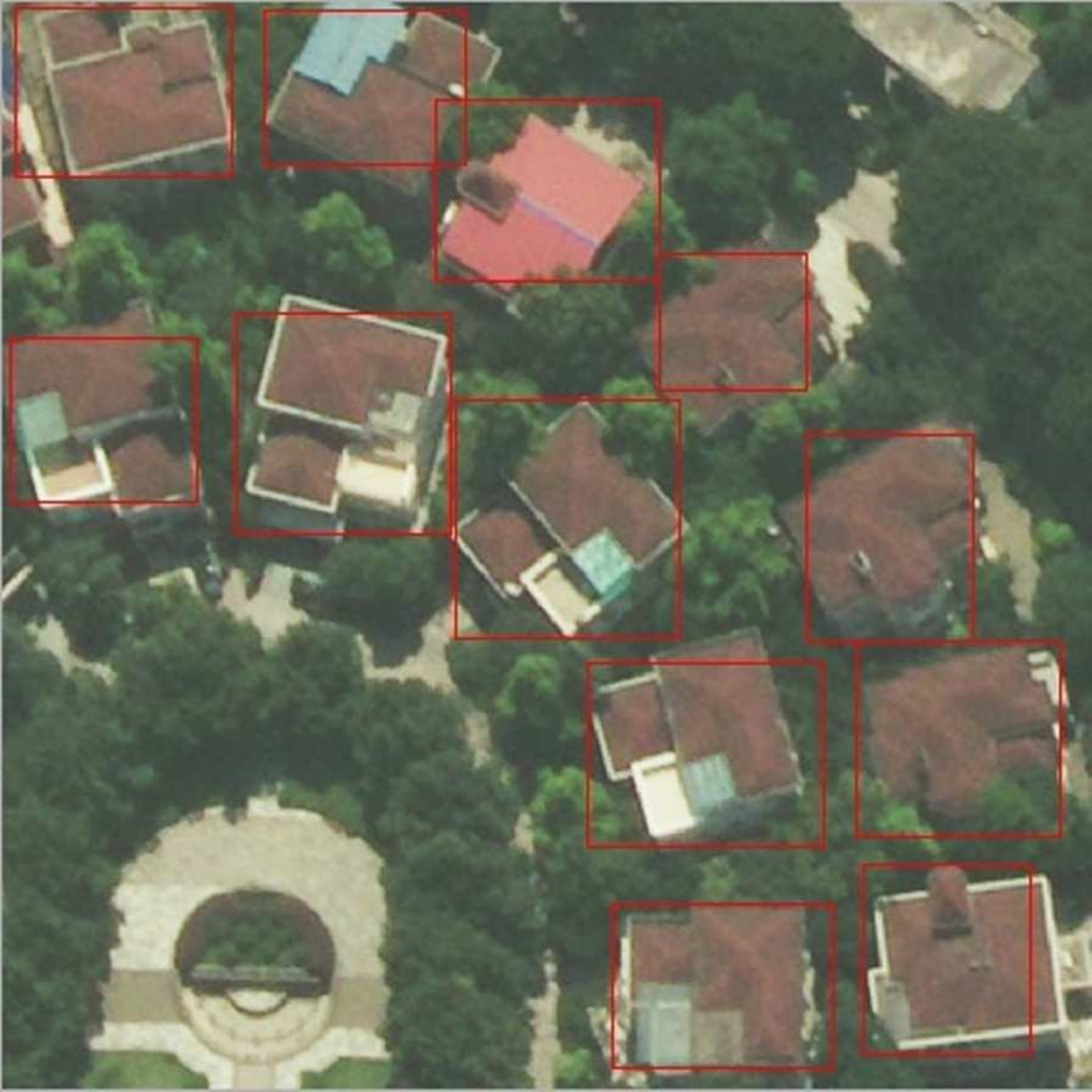}
\expfirstcmp{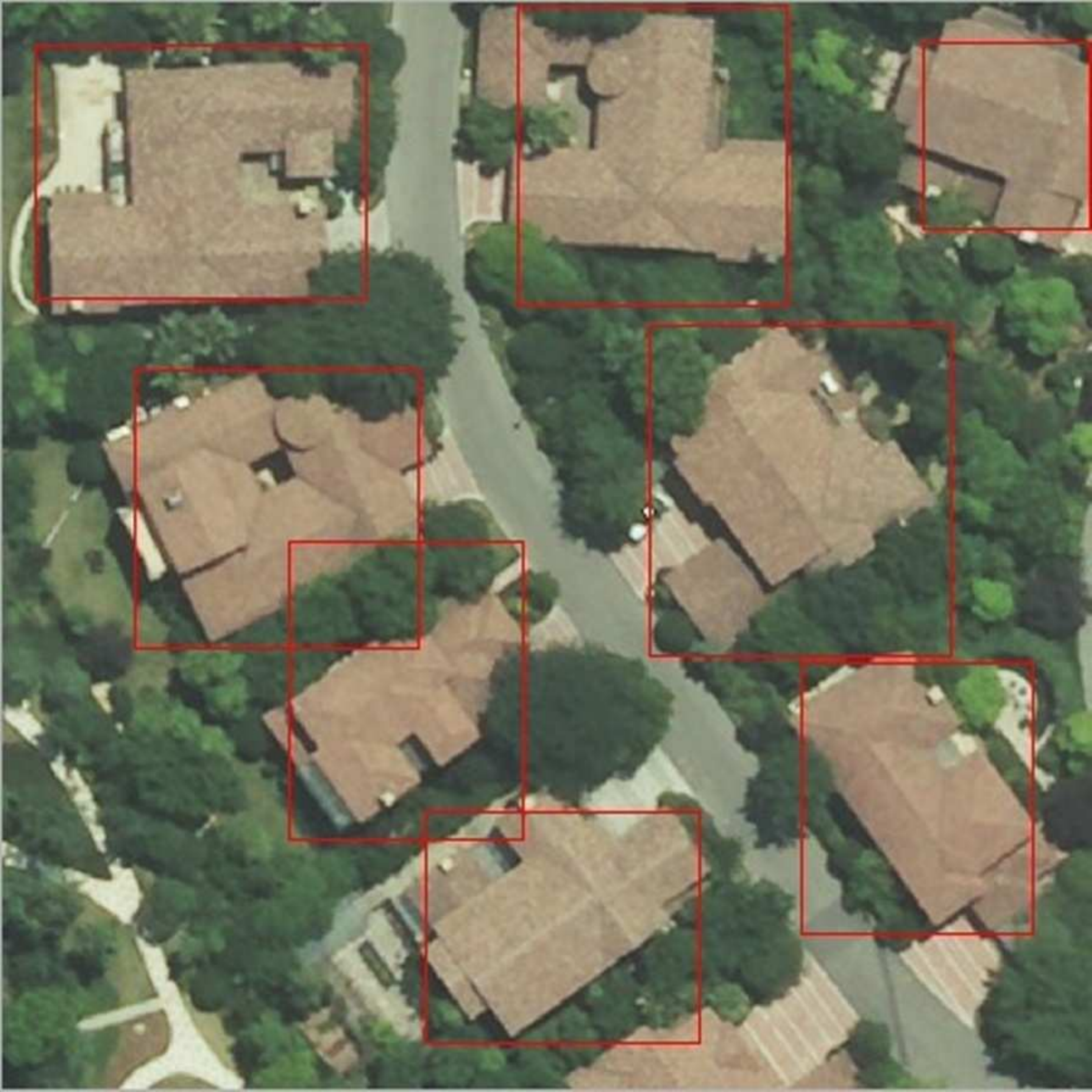}
\\
\expfirstcmp{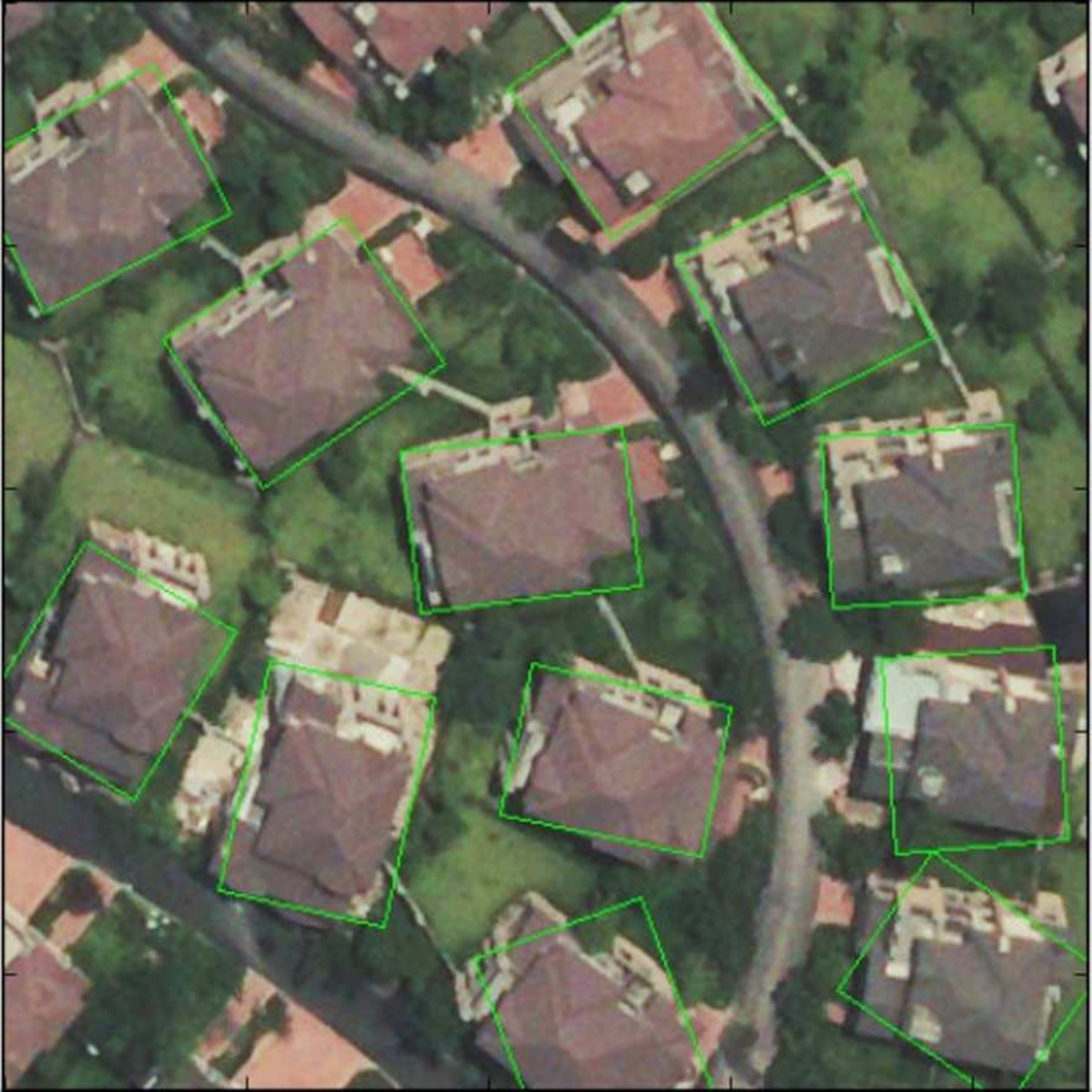}
\expfirstcmp{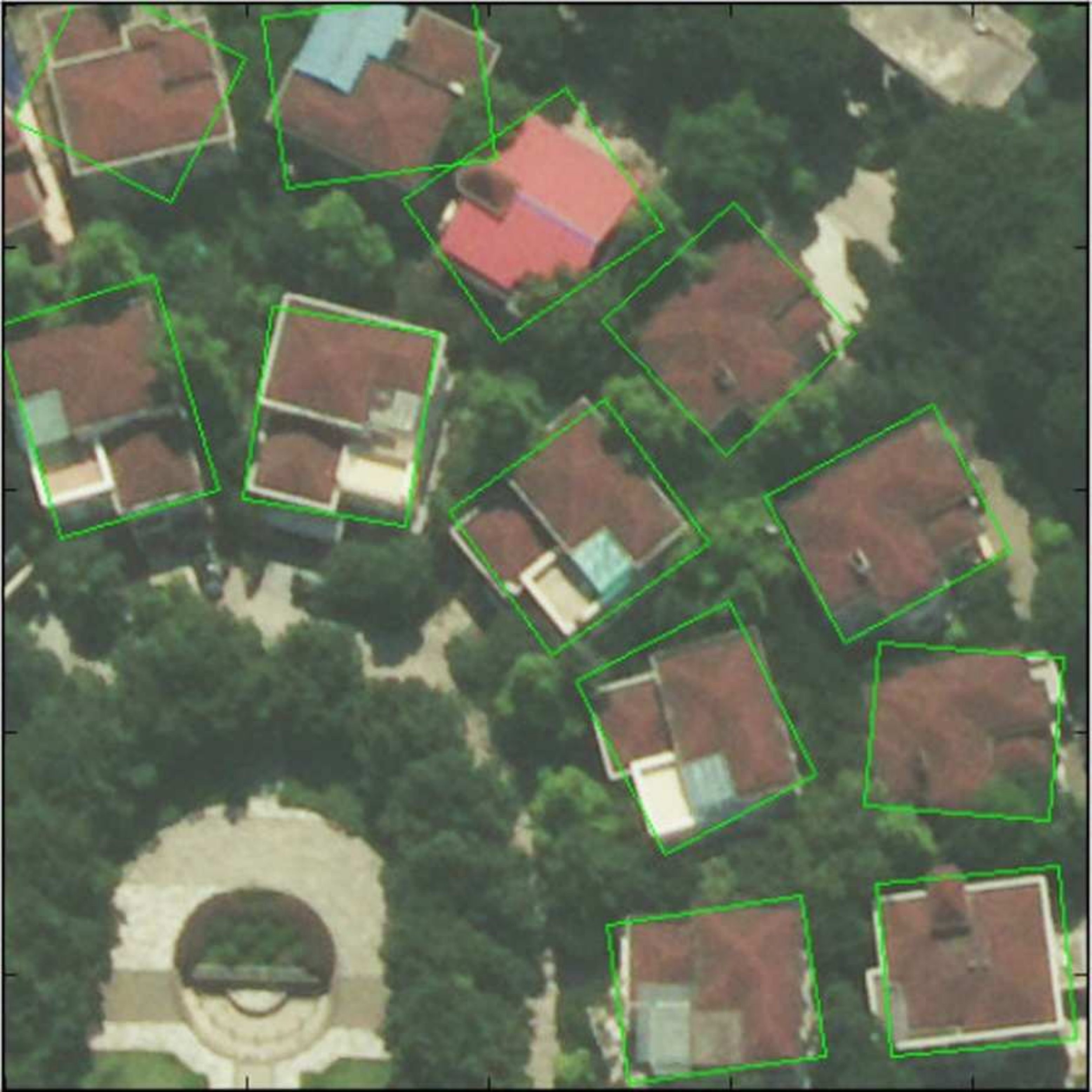}
\expfirstcmp{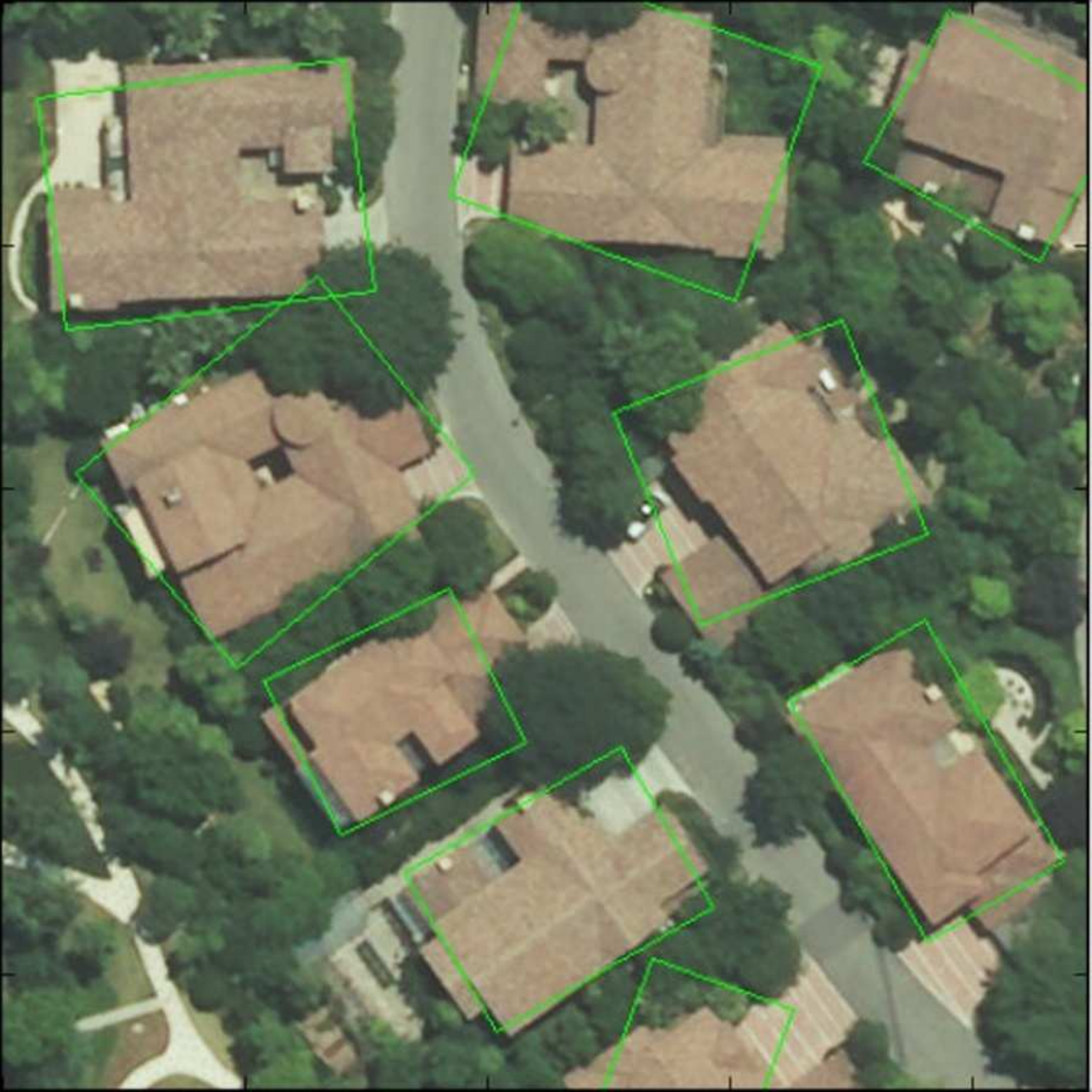}
 	\caption{ Detection results comparison between the axis-aligned
 		(upper row) and our proposed method (bottom row).}
 	\label{fig:448-comparison}
\end{figure}

Intuitive comparisons of the two methods are showed in \figurename{\ref{fig:intuitive comparison}}
 and \figurename{\ref{fig:448-comparison}}.
   Because of the introduction of the angle
regression, our method theoretically outputs tighter and
precise bounding boxes. The proposed method also acquires
ability to better estimate information of objects near the boundary;
and notably, as \figurename{\ref{fig:intuitive comparison}} shows,
 while there is only one building
target in the left-bottom corner being detected by the axis-aligned,
the proposed method detects two and separates them precisely as
the model is enable to learn a joint distribution 
 which allows getting optimal shape of bounding
box according to the predicted orientation.

\begin{figure} 
	\centering 	
\expsecdcmp{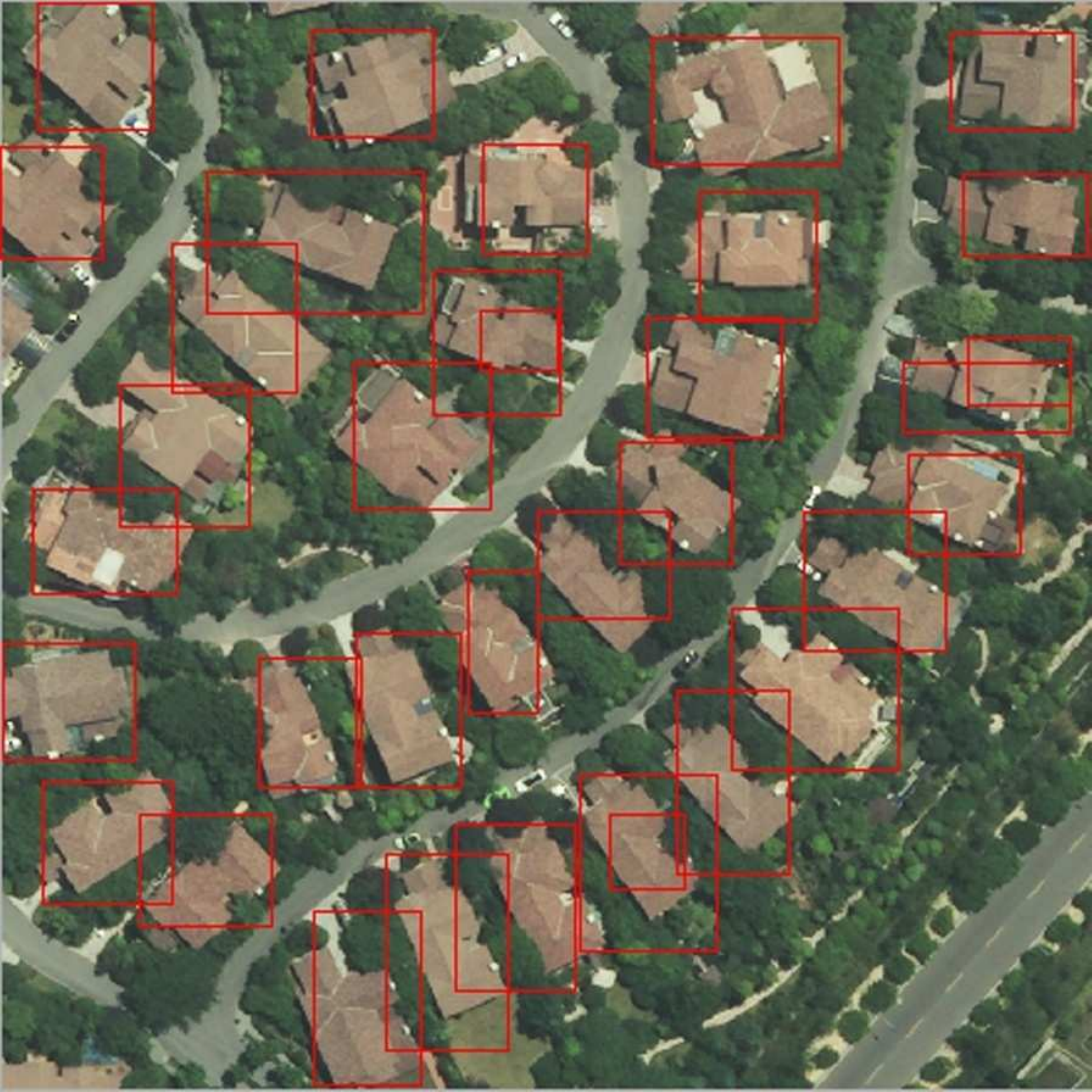}
\expsecdcmp{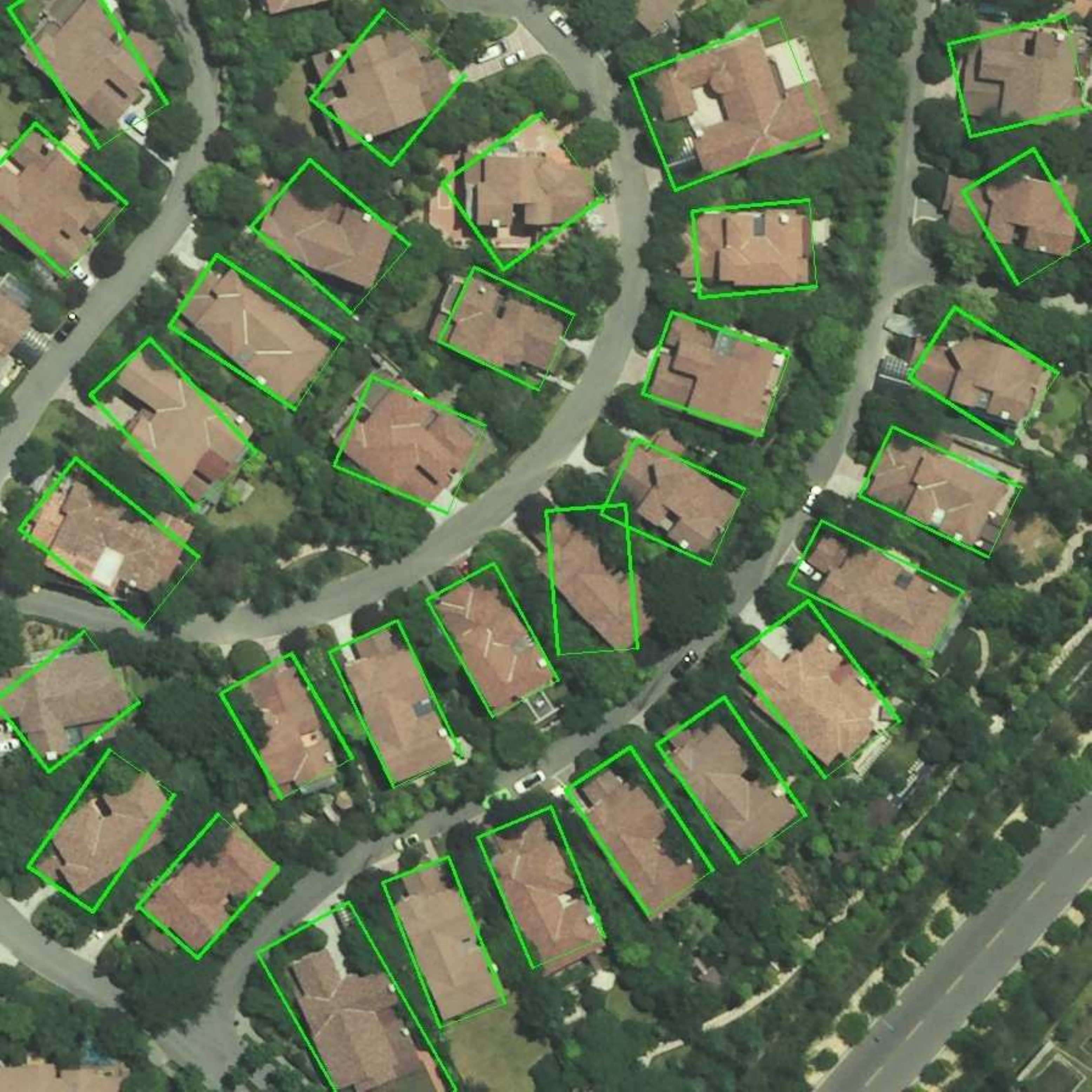}
\\
\expsecdcmp{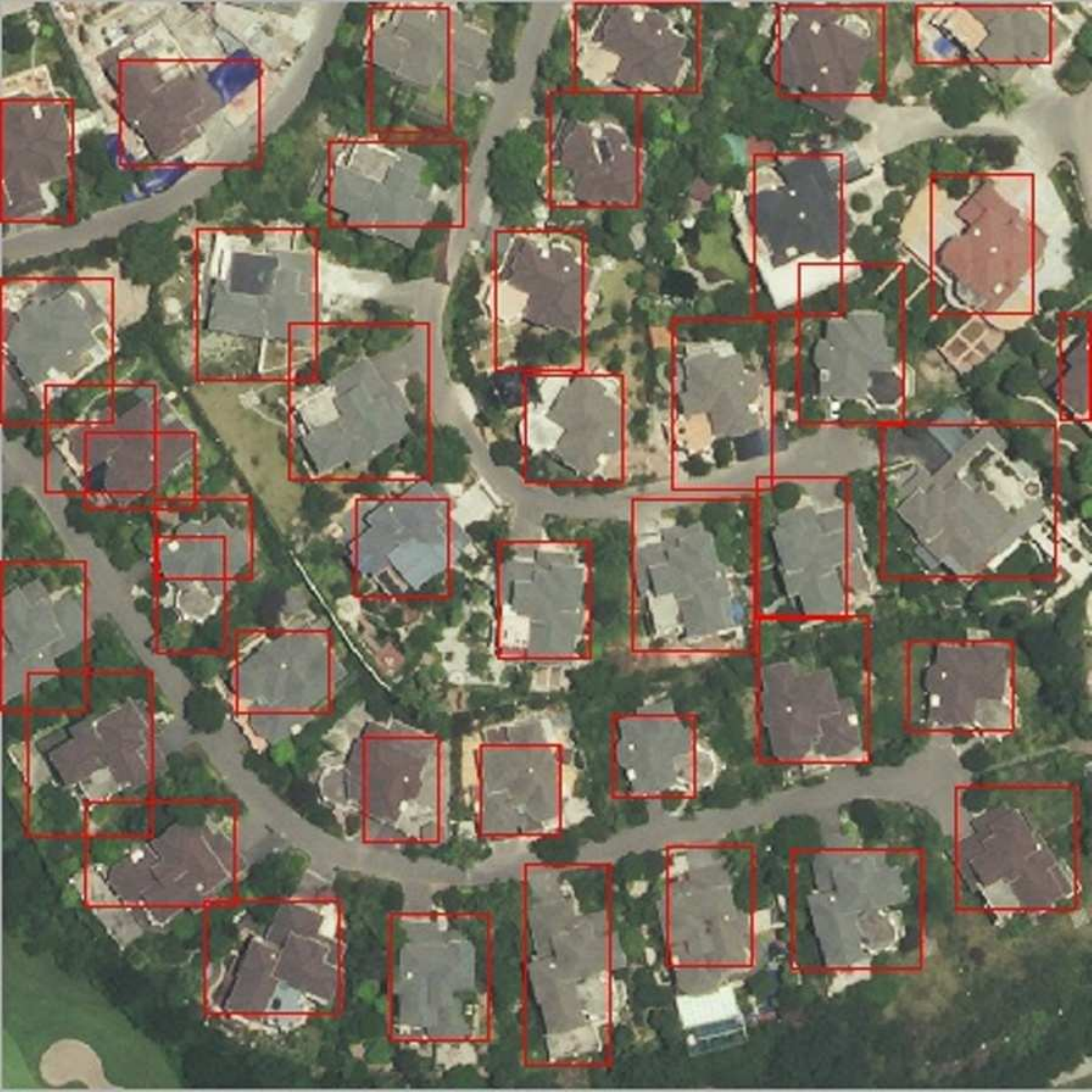}
\expsecdcmp{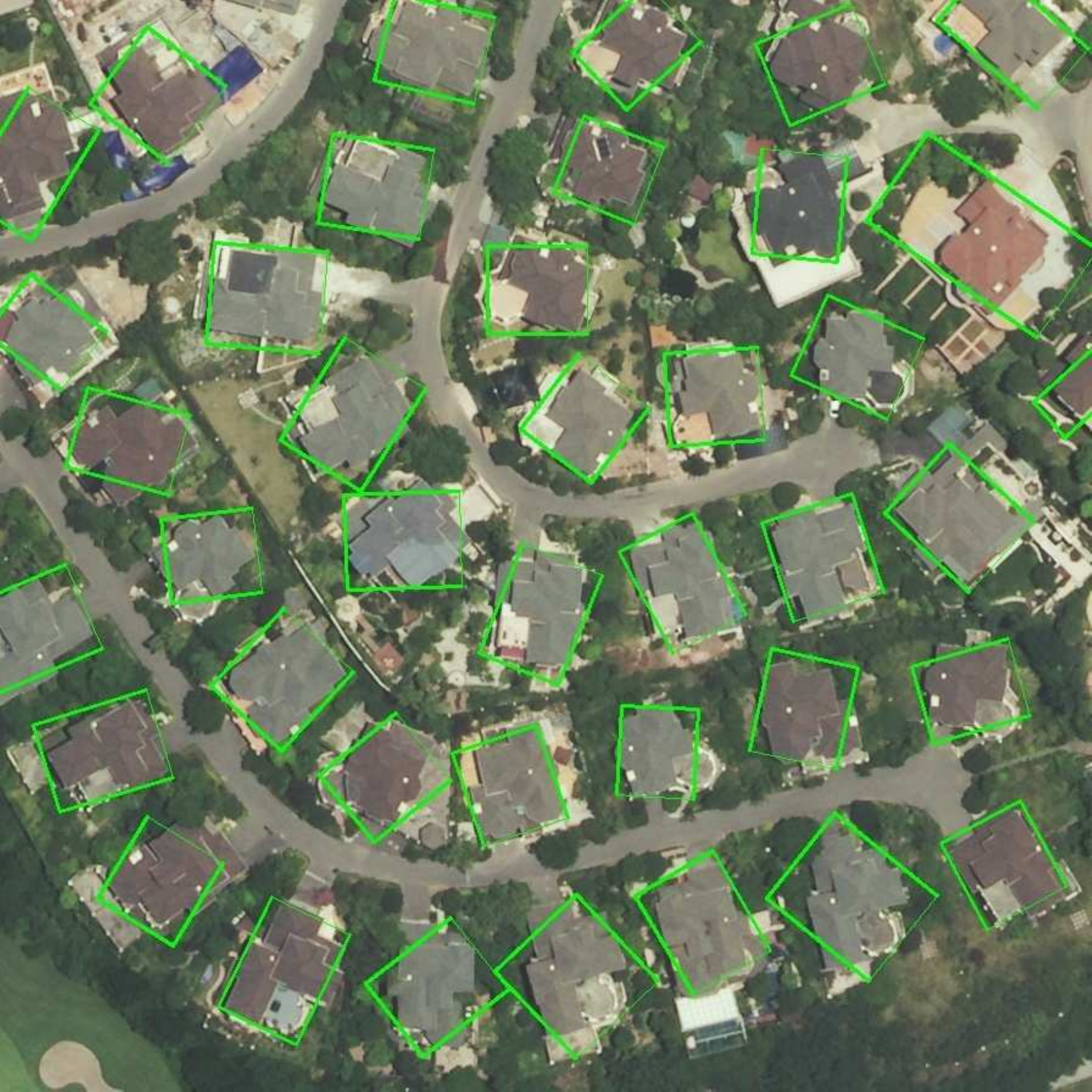}
	\caption{ Detection result on 896$\times$896 images. Contrary to the
		axis-aligned method (left column), our method (right column)
		keeps better visualization performance for large images.}
	\label{fig:896-comparison}
\end{figure}

\begin{figure}
\centering
	\begin{minipage}{0.2\textwidth}
		\centering
		\includeImgPrefix 
		{height=1.5\textwidth}{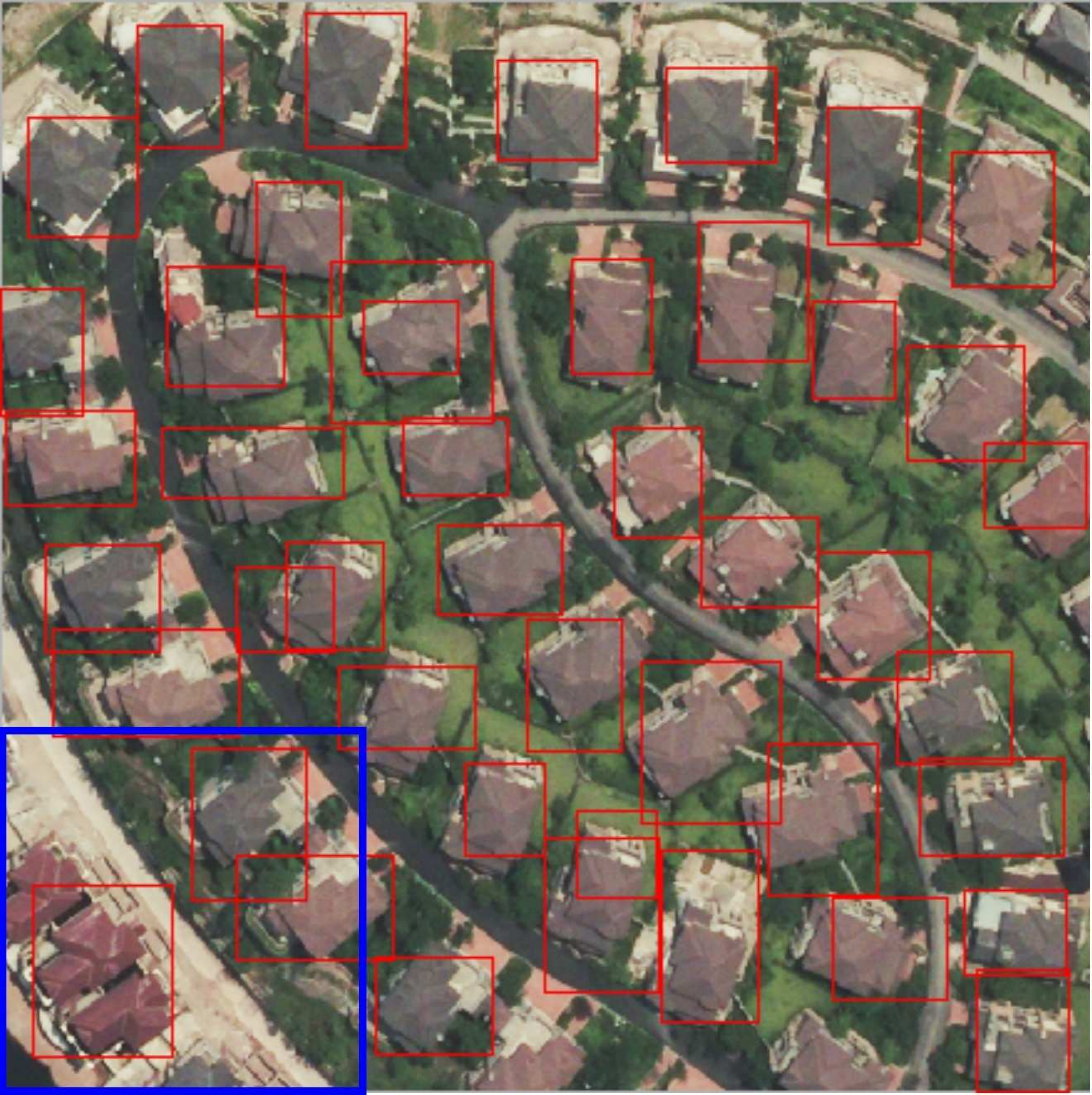}\\
		\hfil(a)\hfil
	\end{minipage}
	\hspace{12mm}
	\begin{minipage}{0.2\textwidth}
	\begin{minipage}{\textwidth}
		\centering
		\includeImgPrefix
		{height=0.678\textwidth}{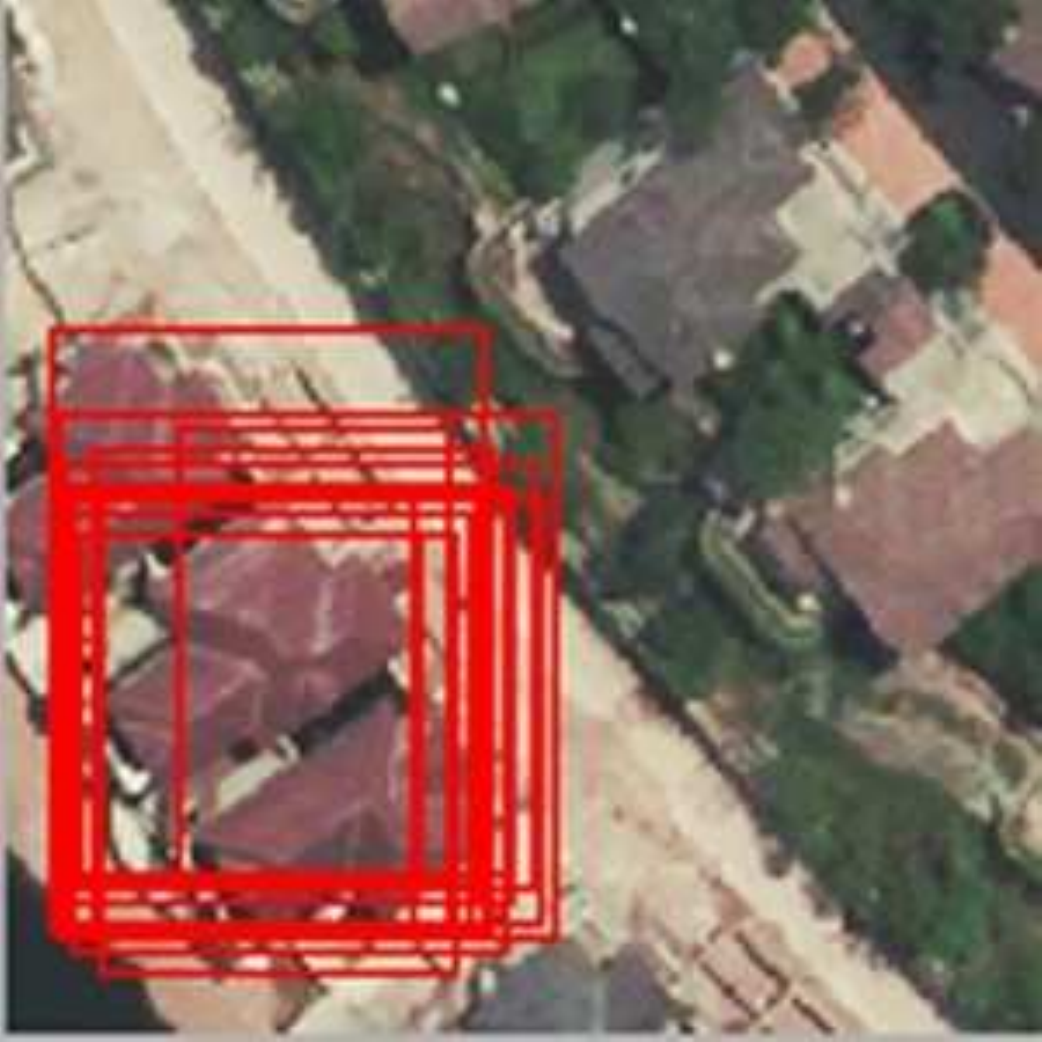}\\
		(b)
	\end{minipage}
		\begin{minipage}{\textwidth}
		\centering
		\includeImgPrefix
		{height=0.678\textwidth}{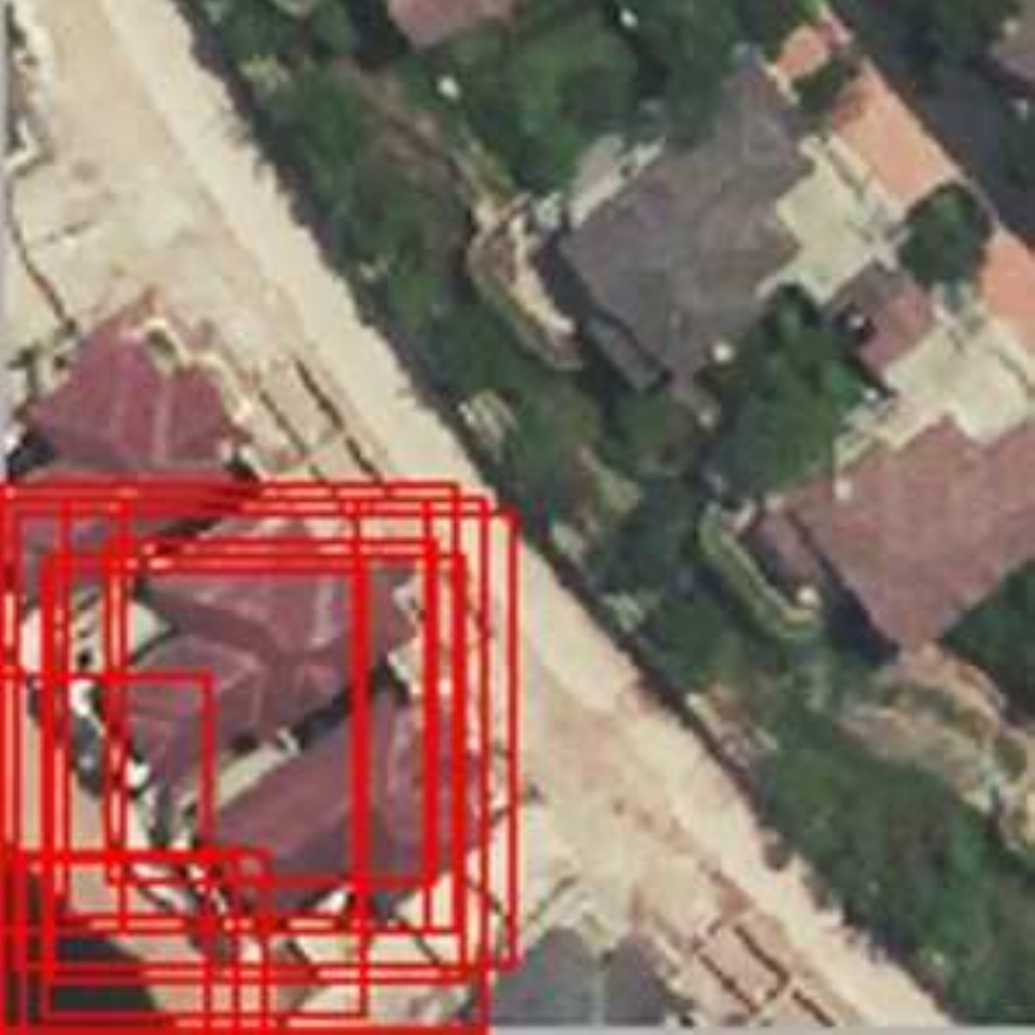}\\
		(c)
	\end{minipage}
	\end{minipage}
	\caption{Detection result in different stages of the axis-aligned
		method. The final result (a) reports only one building object in
		the left-bottom region (denoted by blue box), and the related
		output without NMS (b) clearly shows that the model coarsely
		encloses the two buildings as one unit, similar to the region
		proposals (c) produced by the RPN.}
	\label{fig:896-analysis}
\end{figure}

We extend our experiment to images containing more dense regions 
to get a more comprehensive comparison. The results are
showed in \figurename{\ref{fig:intuitive comparison}}
 and \figurename{\ref{fig:896-comparison}}.
  Although the images tend to contain more objects
and present a more complex environment for detection task, our
method still obtains a better visualization performance over its
counterpart. Compared to the axis-aligned method, the proposed
method shows more advanced detection in handling dense
regions where buildings, usually, with orientation, are
closer to each other increasing the difficulty of detecting.
 In the axis-aligned method, the model is limited to
predict coarse location for target with orientation due to the lack
of orientation regression component. Consequently, the method
is more likely to suffer from dense building regions by
roughly clustering them together.
 As \figurename{\ref{fig:896-analysis}} shows, the
axis-aligned method inappropriately treats the two building objects
as one in the region marked by blue box.
 To study this case, we turn
to the output of the network, as \mbox{( \figurename{\ref{fig:896-analysis}}(b) )} shows,
 which is free from
\textit{NMS} and observe that the two buildings have already been
mixed up. We then, further investigate the region
proposals of the \textit{RPN}. As \figurename{\ref{fig:896-analysis}}(c) shows,
 the RPN produces similar response to that region. In both of the two
stages, the model puts more detection boxes around the center of
the two building objects spanning exactly to the boundary of them two
indicating the model takes the two as a single object.

Meanwhile, as \figurename{\ref{fig:896-comparison}}
 and \figurename{\ref{fig:896-analysis}} show, we notice that the
axis-aligned method is more likely to produce redundant
detections. This is not only caused by inappropriate threshold of
the NMS but the lack of orientation regression as we also observe
there is a considerable amount of overlaps between object boxes
in dense building regions which, if lower IOU threshold in NMS, tends
to result in miss detection.

As the detection result shown in 
\figurename{\ref{fig:intuitive comparison}}(b),
 the convolutional
layers have the ability to preserve orientation information. And, by
introducing orientation regression loss we force the model to
learn to acquire it.

\begin{table} 
	\centering
	\caption{Performance comparison of different
		method/hyperparameter in term of mAP values.}
	\label{tab:mAP-comparison}
	\begin{tabular}{ c  c  c  c }
		\toprule 
		{Method} &  Scales of Anchor  & Angles of Anchor ($\deg$) &mAP \\
		\midrule 
		AA & $64^2,128^2,256^2$ & - & 0.68\\
		AG & $60^2,90^2,130^2$ & -60,0,60 & 0.93\\
		AG-1 & $64^2, 128^2, 256^2$ & -60,0,60 & 0.81\\
		AG-2 & $64^2, 128^2, 256^2$ & 0 & 0.74\\
		AG-3 & $60^2, 90^2, 130^2$ & 0 & 0.77\\
		R2CNN & - & - & 0.94\\
		\bottomrule 
	\end{tabular}
\end{table}

 \begin{figure}
 	\centering
 	\includeImgPrefix{width=0.5\textwidth}
 	{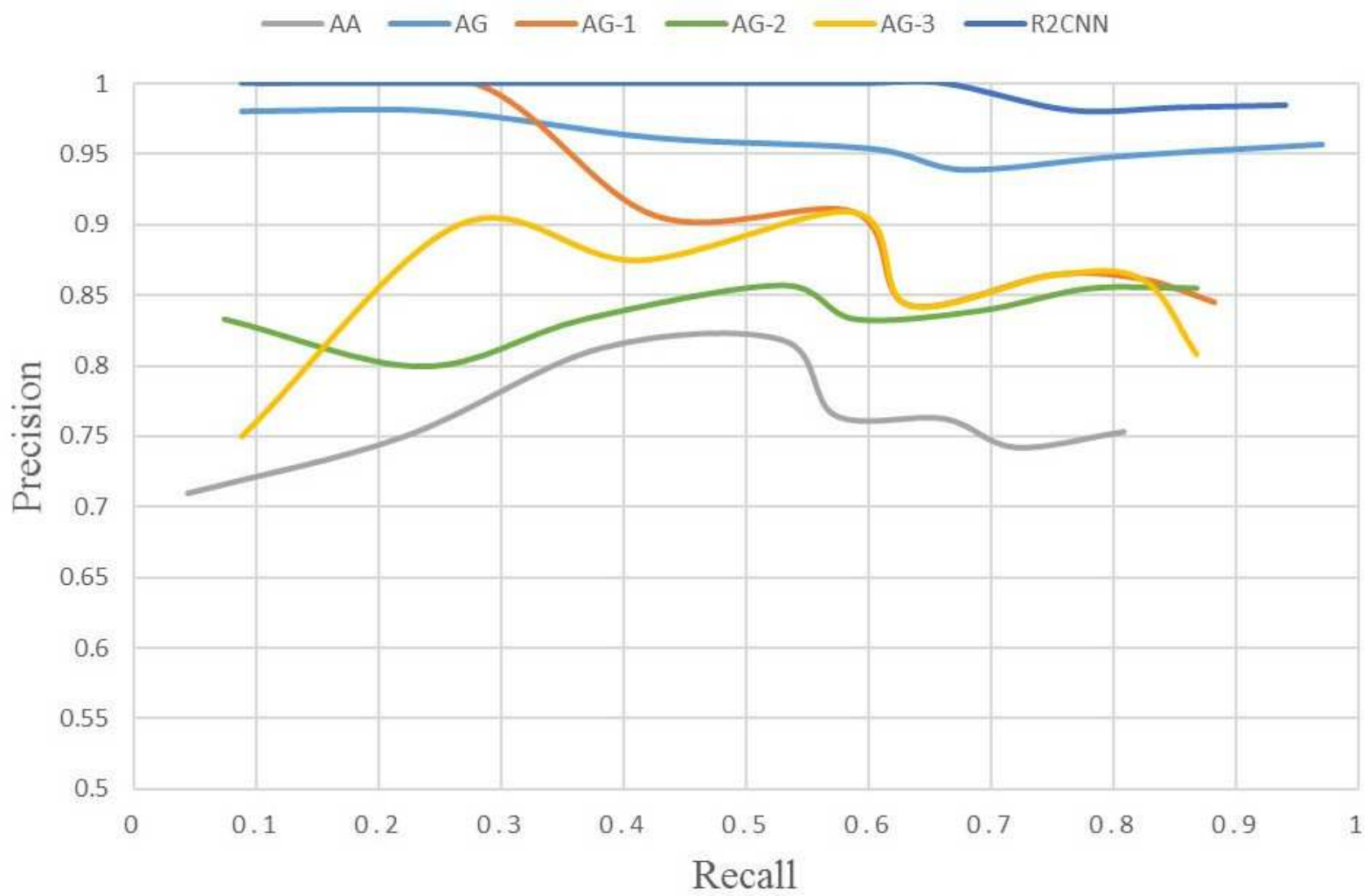}
 	\caption{Precision-recall curves (PRCs). We compare our method
 		under different settings with the axis-aligned method.}
 	\label{fig:PRC}
 \end{figure}

We train our model with different anchor’s settings 
 and evaluate the performance using the two
measures. As \tablename{\ref{tab:mAP-comparison}}
 and \figurename{\ref{fig:PRC}} show, our method achieves
significant improvement compared with the axis-aligned method.
We own this to the additional orientation regression, because of
which the model is able to simultaneously predict the orientation,
location and size. And the higher utilization of bounding box
with respect to IOU makes it more likely to obtain better
performance in terms of mAP value.

%% file: experiments/subsec5.tex
\subsection{Comparison with Other Similar Method}

In parallel with our work, there are some researches which also 
target the problem of orientation not for building in remote sensing
but  text in natural scene.
To have a comparison with the similar work, we
transfer R2CNN model in \cite{jiang_r2cnn:_2017} from 
text detection to our dataset.
and evaluate the two methods on the data.

\begin{figure} 
\centering 	
\begin{minipage}[b]{0.23\textwidth}
 		\centering
\includeImgPrefix{scale=0.28}{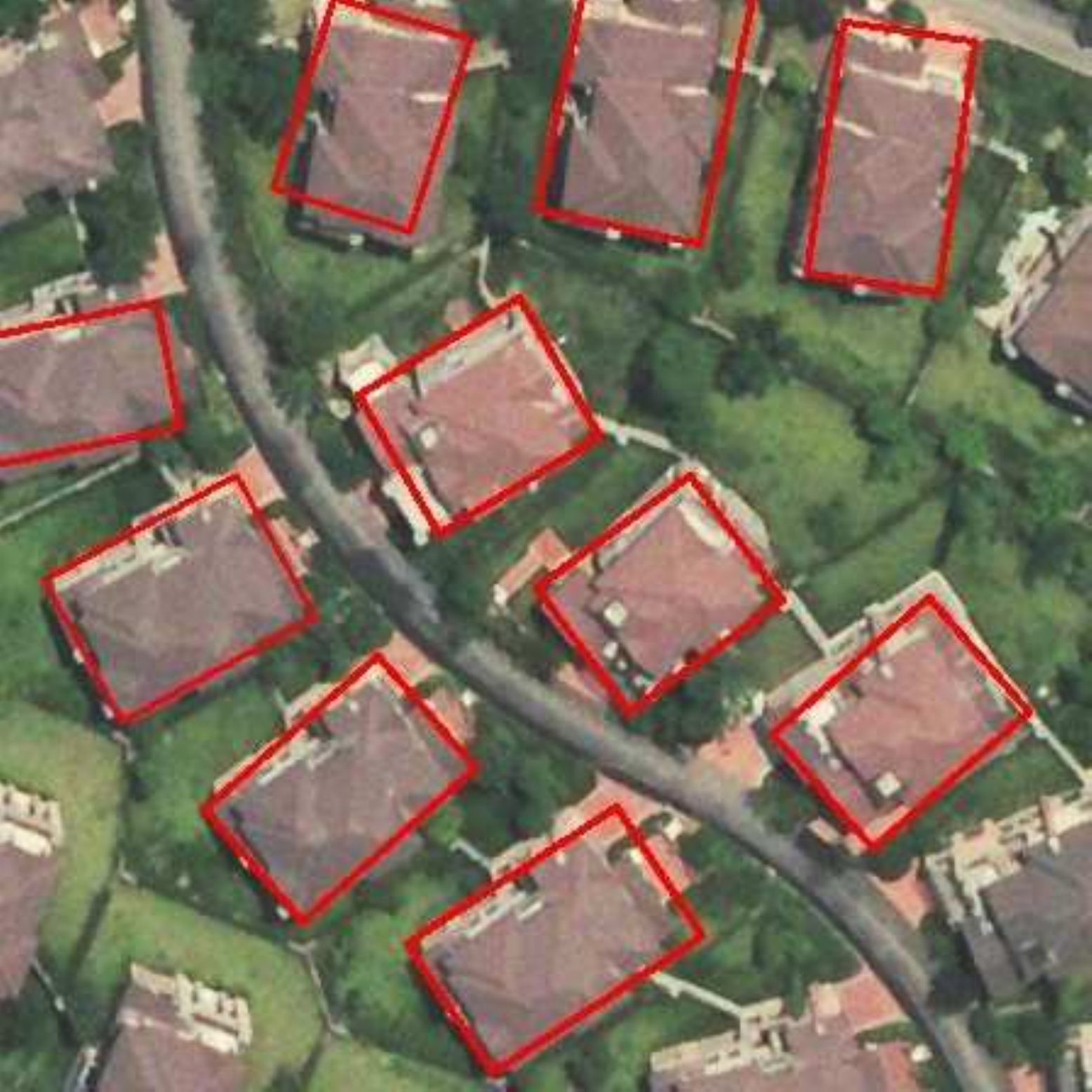}
\end{minipage}
~
\begin{minipage}[b]{0.23\textwidth}
	\centering
	\includeImgPrefix {scale=0.28}{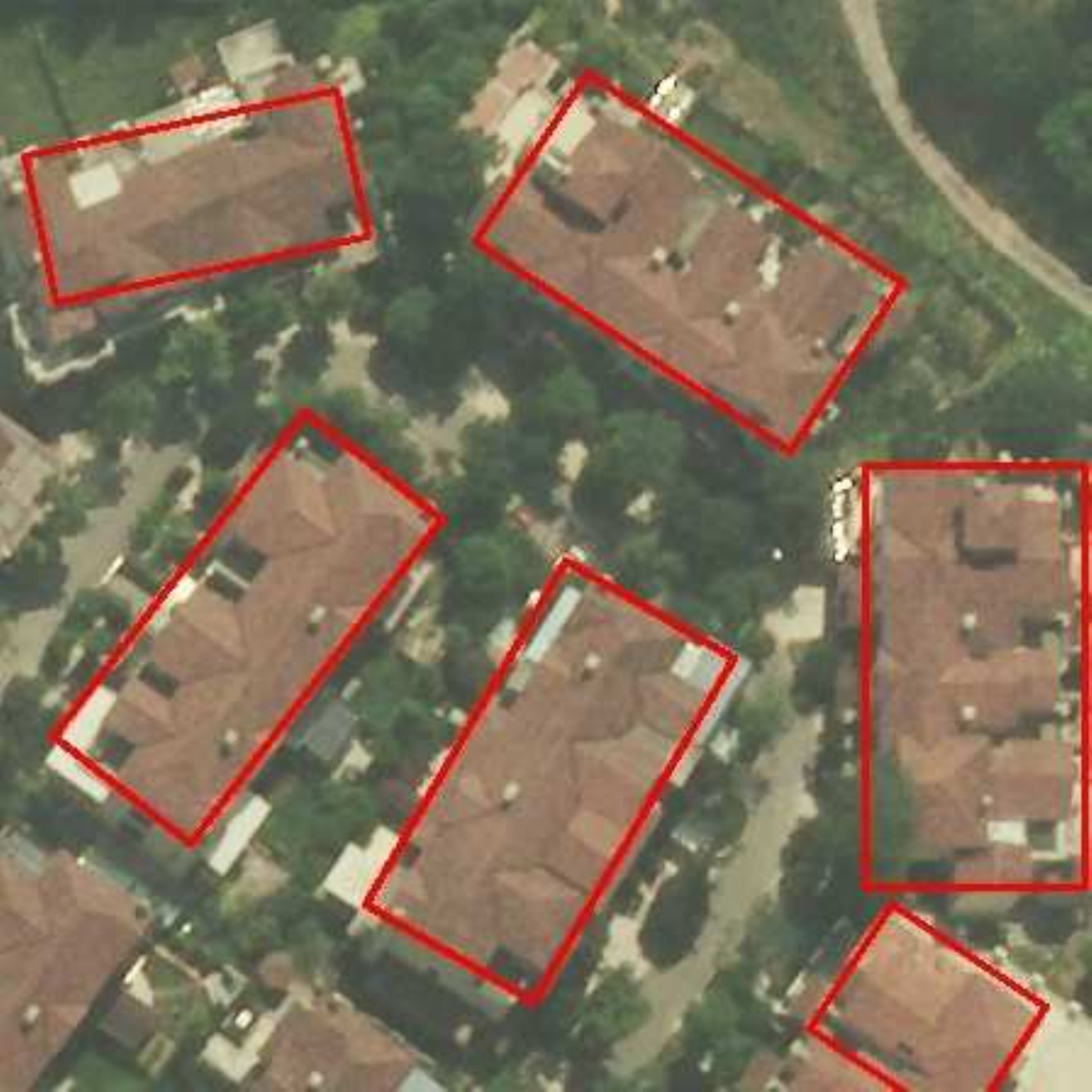}
\end{minipage}
~
\begin{minipage}[b]{0.23\textwidth}
	\centering
	\includeImgPrefix {scale=0.28}{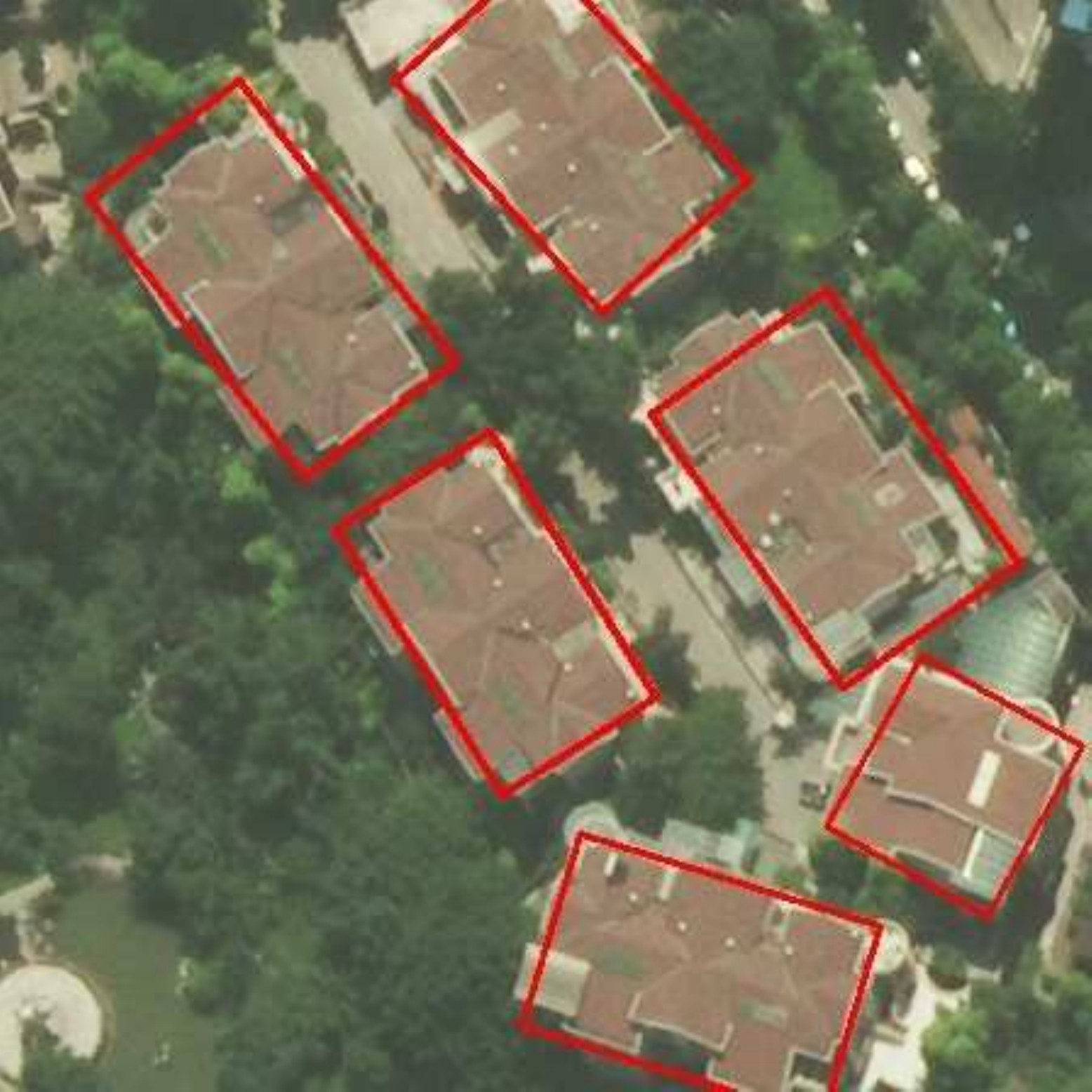}
\end{minipage}

\vspace{1.2mm}

\begin{minipage}[b]{0.23\textwidth}
	\centering
	\includeImgPrefix {scale=0.28}{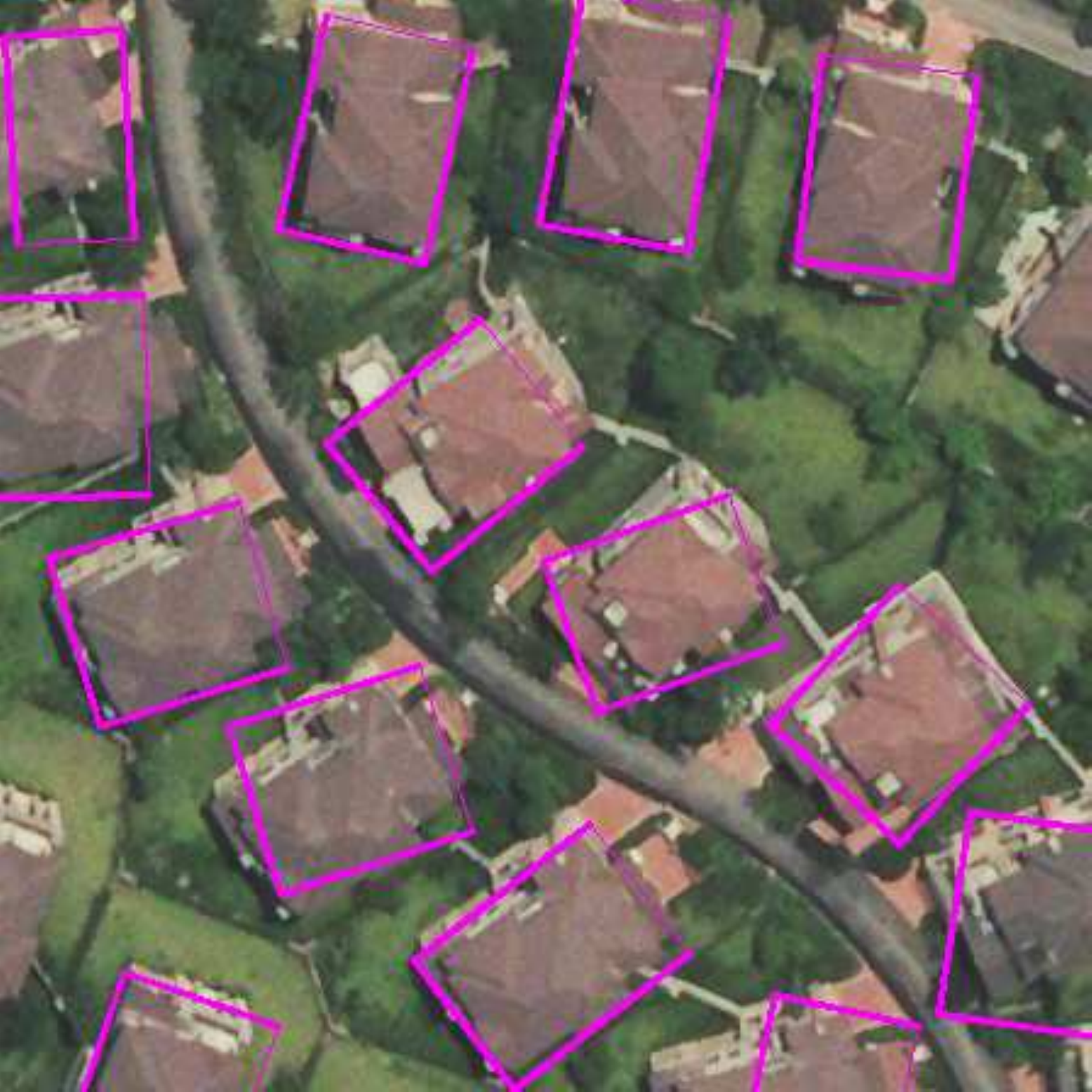}
\end{minipage}
~
\begin{minipage}[b]{0.23\textwidth}
	\centering
	\includeImgPrefix{scale=0.28}{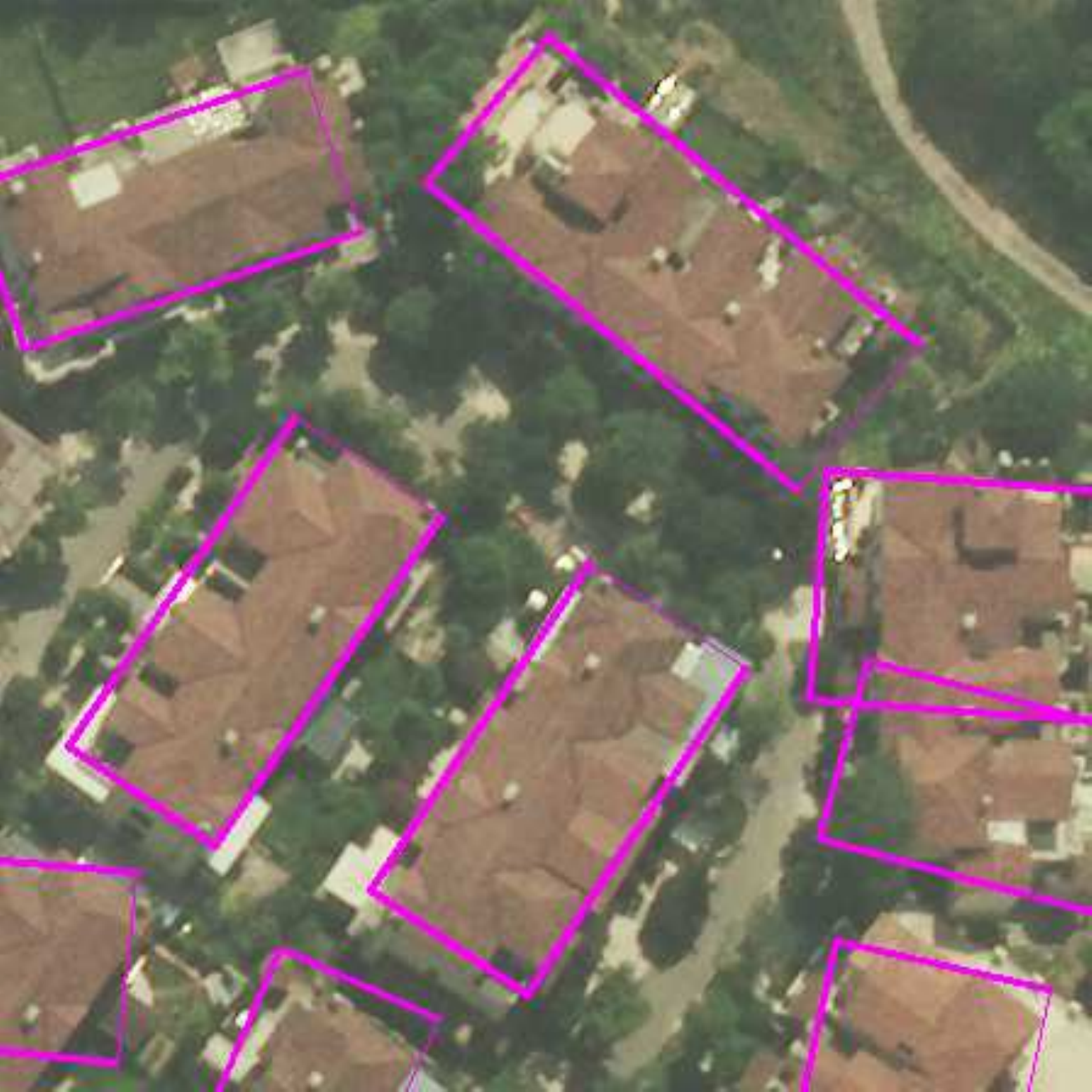}
\end{minipage}
~
\begin{minipage}[b]{0.23\textwidth}
	\centering
	\includegraphics[scale=0.28]{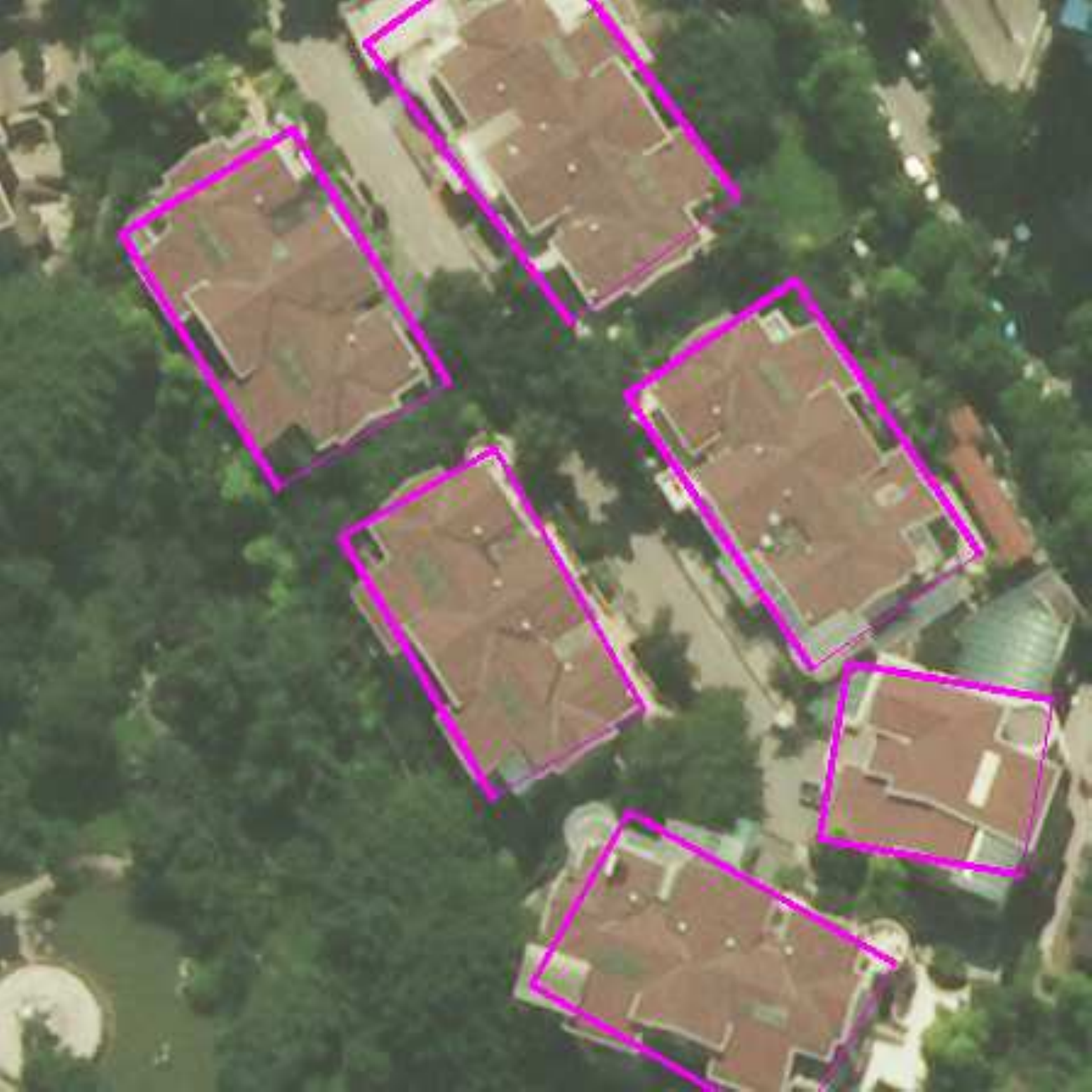}
\end{minipage}
\caption{Detection results of R2CNN method (upper row) and ours (bottom).} 	
 	\label{fig:comparison-AGvsR2CNN}
\end{figure}

\figurename{\ref{fig:comparison-AGvsR2CNN}} shows detection 
results for the methods.
Both of the methods show similar performances in locating
buildings and estimating orientation.
In detail, R2CNN tends to obtain more precise detection as
it takes two-stage framework for refining prediction.
However, our method is more relying on local information
since the model uses sliding kernel to regress the boxes, and
as a result our method tends to discover more buildings in image margins
which can be considered as partial occlusion.

The performance of R2CNN is measured in the last row of \tablename{\ref{tab:mAP-comparison}}
and the PR curve is drawn in \figurename{\ref{fig:PRC}}. As the results show
the two methods are almost neck and neck in terms of mAP. Although R2CNN obtains
a slightly better score, ours model takes a more unified framework and is more compact.

In \tablename{\ref{tab:mAP-comparison}}, we use different settings to explore the validity
of the proposed detection model as well as multiangle anchor.
Comparison between \textit{AA} and \textit{AG-2} directly 
demonstrates the   value of detecting with orientation as 
the settings of \textit{0 deg} for the angle of the anchors mounts
to generating identical anchors for the two methods.

\textit{AG} with \textit{AG-3} aims to test the validity of the proposed 
multiangle anchors. The higher score achieved by the former
suggests that with properly adding angles of the proposed orientation anchors detection
would have a significant improvement.

%% file: conclusion/conclusion.tex
\section{Conclusion}\label{sec:conclusion}

In this work, we aim to bridge the gap between the popular CNN-based
 axis-aligned object detection method and the orientation
variation of buildings that generally exists in remote sensing
images. To tackle this, we introduce orientated anchors and
additional dimensionality to regress orientation for each building
object. To address the problem of calculating IoU overlap
between any two arbitrarily oriented boxes, we develop an
algorithm for estimating which is feasible to implement in GPU
for fast computing. To train and test our model, we construct a
dataset of remote sensing images and manually label building
objects. Our method is implemented in an \mbox{end-to-end} fashion and
tends to produce tighter bounding boxes of higher IoU overlap
with building objects which leads to better performance over the
axis-aligned one in terms of mAP. 
Also, we compare our method with the related work proposed fro rotational text region detection.
They achieve closed scores but differ at  structure.
Our model presents an unified and compact framework, however it
errors of orientation estimation
 and shows sensitive to
hyperparameter settings.
Hence, we will take more efforts on obtaining more precise orientation
estimation in our future study.